\documentclass{egpubl}
\usepackage{ICAT-EGVE2026}

\WsSubmissionJoint 

\usepackage[T1]{fontenc}
\usepackage{dfadobe}  
\usepackage{booktabs}
\usepackage{multirow}

\usepackage{cite}  
\BibtexOrBiblatex
\electronicVersion
\PrintedOrElectronic
\ifpdf \usepackage[pdftex]{graphicx} \pdfcompresslevel=9
\else \usepackage[dvips]{graphicx} \fi

\usepackage{egweblnk} 

\usepackage{cite}
\usepackage{amsmath,amssymb,amsfonts}
\usepackage{algorithmic}
\usepackage{graphicx}
\usepackage{textcomp}
\usepackage{xcolor}
\usepackage{cite}
\usepackage{subfigure}
\usepackage{url}
\usepackage{hyperref}
\usepackage{svg}
\def\BibTeX{{\rm B\kern-.05em{\sc i\kern-.025em b}\kern-.08em
    T\kern-.1667em\lower.7ex\hbox{E}\kern-.125emX}}

\usepackage{microtype}

\title[ERUPT]%
      {ERUPT: An Open Toolkit for Interfacing with Robot Motion Planners in Extended Reality}

\author[I. Ngui et al.]
{\parbox{\textwidth}{\centering I. Ngui$^{1}$\orcid{0009-0009-3580-0220},
        C. McBeth$^{1}$\orcid{0000-0002-6826-0349},
        A. Santos$^{2}$,
        G. He$^{1}$,
        K. Mimnaugh$^{1}$\orcid{0000-0002-6306-1674},
        J.\,D. Motes$^{1}$\orcid{0000-0002-9553-7331},
        L. Soares$^{2}$\orcid{0000-0003-3187-4168},
        M. Morales$^{1,3}$\orcid{0000-0003-1824-2350}
        and N.\,M. Amato$^{1}$\orcid{0000-0001-5817-5290}
        }
        \\
{\parbox{\textwidth}{\centering $^1$University of Illinois Urbana-Champaign, United States\\
         $^2$Insper, Brazil \\
         $^3$Instituto Tecnológico Autónomo de México, México
       }
}
}

\begin{document}


\maketitle
\begin{abstract}
We present the Extended Reality Universal Planning Toolkit (ERUPT), an extended reality (XR) system for interactive motion planning.
This paper serves to introduce our open-source system to others who can use it as a base to develop immersive robot interaction applications. 
Our system allows users to create and dynamically reconfigure environments while they plan robot paths.
ERUPT uses XR to provide a broad range of natural interaction capabilities, allowing users to grab and adjust objects in the environment similar to interaction in the real world, rather than using a mouse and keyboard with the scene projected onto a 2D computer screen.
Our system integrates with MoveIt, a manipulation planning framework, allowing users to send motion planning requests and visualize the resulting robot paths in virtual or augmented reality.
We provide a broad range of interaction modalities, allowing users to modify objects in the environment and interact with a virtual robot.
Our system allows operators to visualize robot motions, ensuring desired behavior as it moves throughout the environment, without risk of collisions within a virtual space, and to then deploy planned paths on physical robots in the real world.
Code can be found at \url{https://github.com/parasollab/erupt}.


\end{abstract}  
\section{Introduction}

Robots are becoming more prevalent in many aspects of life ranging from industrial settings like factories to domestic assistive scenarios.
These applications often feature users reconfiguring environments, requiring robots to adapt their behavior.
For example, manufacturing engineers may reposition workspaces and safety barriers while designing human-robot collaborative workstations, warehouse operators may rearrange storage locations to improve workflow, and remote supervisors may adjust a robot's operating environment during teleoperation.
In each of these cases, users must modify the environment, reposition robots, and evaluate motion plans to ensure tasks can be completed safely and effectively.

A common way to configure the environment setup and query collision-free robot paths is to use MoveIt~\cite{coleman2014reducing}, a manipulation planning framework integrated with the Robot Operating System (ROS).
Using this approach, environments are typically visualized as projected onto a two-dimensional computer screen, which may limit the operator's spatial understanding, potentially requiring them to view the environment from many angles to accurately perceive the robot's behavior moving throughout the space~\cite{McIntire_Havig_Geiselman_2014}.

Head mounted display (HMD) technology has made it possible to immersively visualize three-dimensional environments, move around within them, and interact with objects. Extended Reality (XR), which encompasses virtual and augmented reality~\cite{Skarbez_Smith_Whitton_2021}, offers a natural modality for reconfiguring environments and visualizing robot motion. Virtual robots are particularly well suited to exploring potential robot behavior because there is no risk of collision with the environment and nearby human operators. XR also offers a broad range of natural interaction capabilities, allowing users to adjust the position and rotation of environment obstacles by manipulating them in three dimensions, which may be impossible in the real world. 

\begin{figure}
    \centering
    \includegraphics[width=0.65\linewidth]{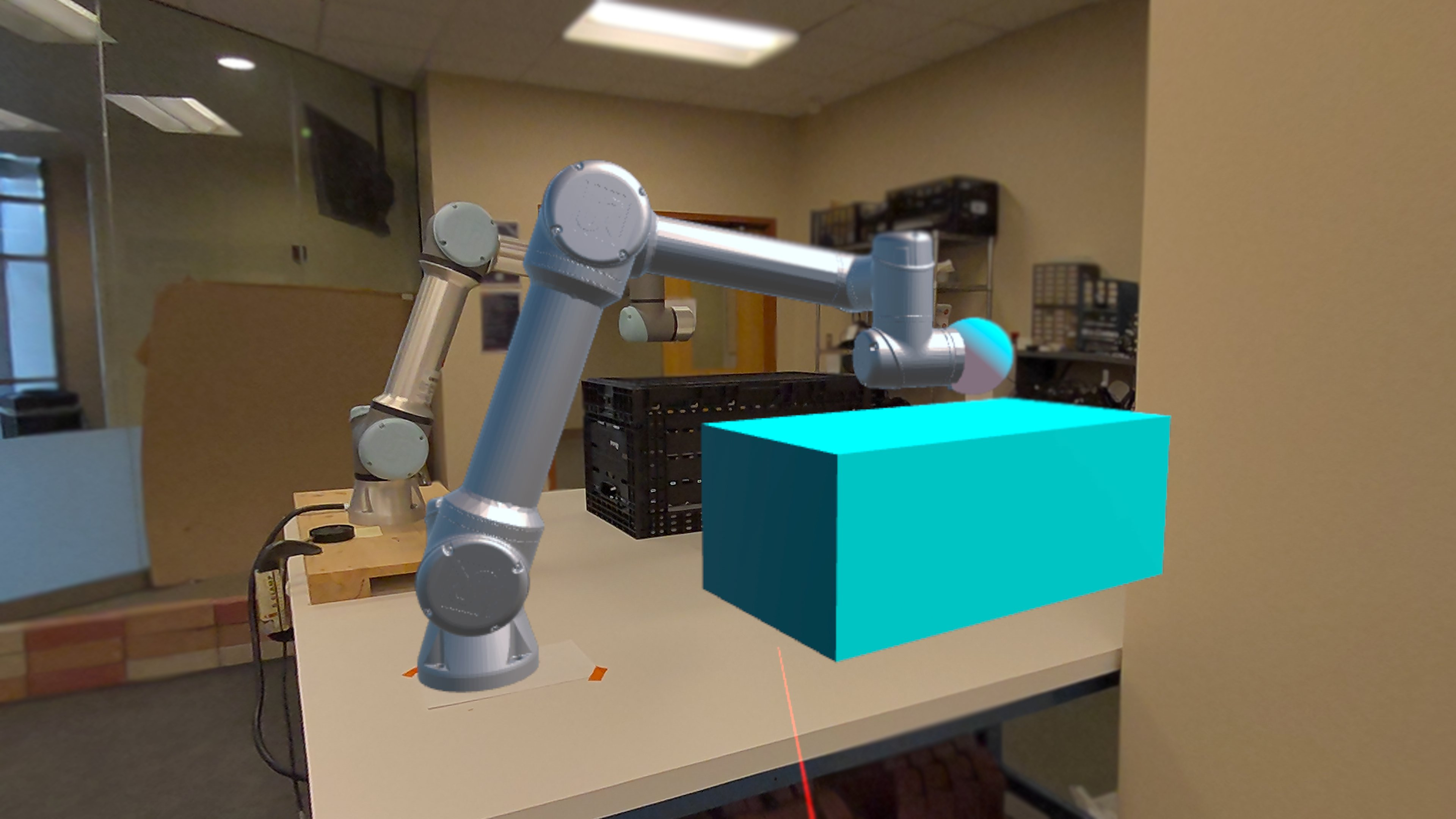}
    \caption{An example environment setup. A virtual robot and obstacle are set to mirror a physical environment so that an operator can prototype obstacle configurations and robot motions.}
    \label{fig:teaser}
\end{figure}

To leverage the advantages provided by immersive technology to enable rapid environment reconfigurations and (re)programming of robot behavior, we propose the Extended Reality Universal Planning Toolkit (ERUPT). ERUPT is an open-source extended reality application that allows operators to import their own robots, reconfigure environments, plan robot paths, and visualize robot motions.
While prior works have explored XR environments for robot programming (Tab.~\ref{tab:landscape}), our system provides a wider range of interaction capabilities for configuring the environment and planner.
Our contribution focuses on the integration and engineering required to support these capabilities and maintain a consistent representation of the robot and environment across the XR application and existing robotics frameworks.
Our system integrates with ROS2, allowing users to easily extend the system for their applications. We provide an interface to the MoveIt planning framework, allowing operators to update the environment representation and plan and visualize paths generated using standard motion planners. 

The larger community of robotics researchers and operators can also expand upon our system's base capabilities to support a broader range of robotics problems and planning frameworks. 
For example, ERUPT may be extended to serve as a platform for collecting demonstrations and human feedback for learning-based planners and preference learning approaches. It may also support immersive remote robot supervision and teleoperation, allowing users to explore new human-robot interaction paradigms while leveraging our visualization and environment modeling infrastructure.
Our contributions are:
\begin{itemize}
    \item An open-source XR toolkit for robot motion planning enabling users to configure robots, goals, and environment objects in an immersive interface
    \item A ROS2 \& MoveIt integration layer that synchronizes XR scene edits with the MoveIt planning scene allowing collision objects created or modified in XR to be used directly by motion planners
    \item An interactive planning workflow for configuring, requesting, inspecting, and executing motion plans including planner selection and preview before physical execution
    \item A quantitative evaluation of system performance, characterizing rendering frame rate under task load as well as communication and scene update latency.
    \item Demonstrations in simulation and on physical robots showing ERUPT's ability to reconfigure scenes, request collision-aware plans, preview generated motion plans, and execute selected trajectories.
\end{itemize}

\begin{figure}
    \centering
    \includegraphics[width=0.47\textwidth]{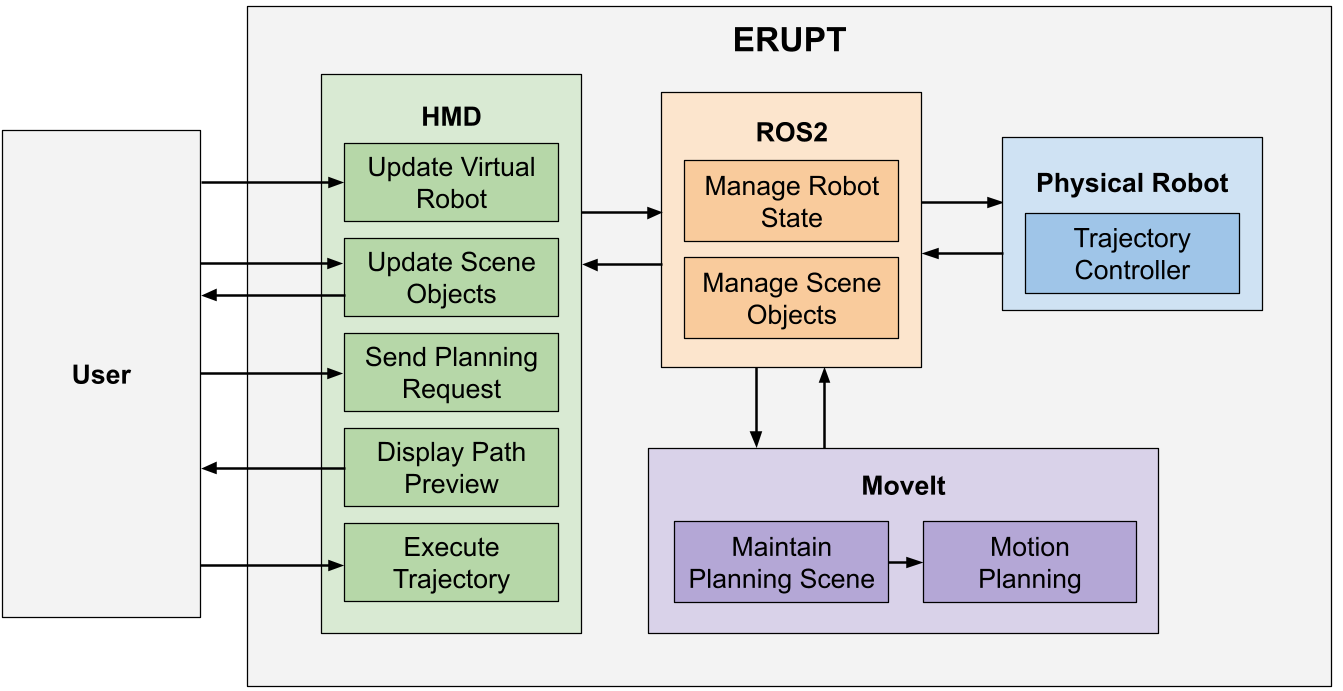}
    \caption{An overview of the ERUPT system. Through the HMD, the user interacts with the virtual environment and robot, which are managed by the game engine. Updates to objects in the environment are synced over ROS2 between the XR representation and the MoveIt planning scene. Users send motion planning requests, which are forwarded to MoveIt, and the resulting paths are visualized to the user. After planning a satisfactory path, the user may execute it on a physical robot.}
    \label{fig:system-diagram}
\end{figure}


\section{Background and Related Work}
\label{sec:background_related_work}

\begin{figure}
    \centering
    \includegraphics[width=0.9\linewidth]{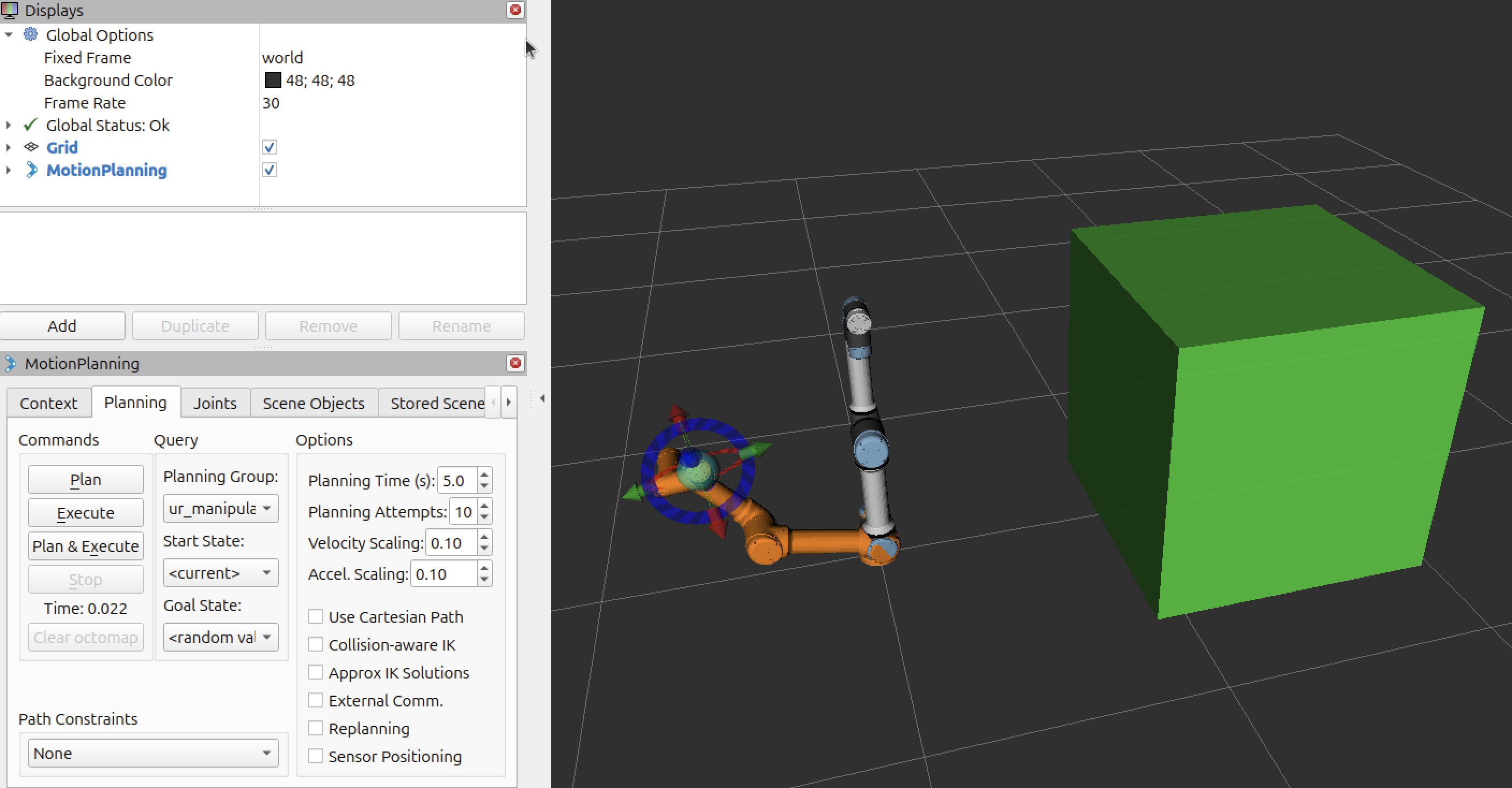}
    \caption{The 2D MoveIt planning interface in RViz.}
    \label{fig:moveit}
\end{figure}

\subsection{ROS, MoveIt, and Unity}
A large contributor to the growing area of robotics is its open-source software environment.
The Robot Operating System (ROS)~\cite{quigley2009ros, ros2} is an open-source set of software libraries and tools for creating and customizing robot applications.
ROS provides interfaces for many different robot components, such as sensors, motion planners, kinematics, and more, letting you use their default components or develop custom components for a specific application.

A popular tool which leverages ROS for manipulation planning is MoveIt~\cite{chitta2012moveit, coleman2014reducing}.
MoveIt is an open-source robotic manipulation platform which leverages ROS to provide manipulation-related robot components such as motion planning, grasping, perception, and others.
In order to enable all these features, MoveIt provides a set of interfaces for each of the manipulation components, which interact with their environment representation, the planning scene.
The planning scene is the environment and robot state representation for MoveIt storing collision objects as well as dynamic environment information, which can be used by the motion planners to compute robot trajectories.
MoveIt provides a set of services for planning which allow users the ability to choose their desired planner as well as to customize the parameters of the planning algorithms to their specific applications which we leverage in our work.

Visualizations of plans and MoveIt components are provided by the Robot Visualizer (RViz; Fig~\ref{fig:moveit}; https://wiki.ros.org/rviz).
RViz is a screen-based robotics visualization tool which displays information about ROS data letting users see sensor, robot, and environment information.
Recently, roboticists have begun exploring the use of game engines such as Unity or Unreal Engine as simulation environments~\cite{9812353, de2019analysis}.
These game engines have also decreased the difficulty of creating XR robotics applications as they support XR capabilities.
Unity (https://unity.com), in particular, has even seen a community of roboticists develop packages to interface with ROS, such as the ROS TCP Connector (https://github.com/Unity-Technologies/ROS-TCP-Connector) based on its predecessor ROS$\#$ (https://github.com/siemens/ros-sharp).
The ROS TCP Connector allows developers to create ROS nodes in Unity which can communicate with pure ROS elsewhere.

\subsection{Extended Reality for Robotics}

In recent years, XR systems have been proposed to address many robotics problems and applications, including teleoperation, robot-assisted surgery, and robot training systems~\cite{suzuki2022ar, makhataeva2020ar}. 
Many works explore human-robot collaborative scenarios, where XR has been leveraged to improve safety.
Some works utilize XR as a means of transparency, providing information to the user such as robot intent~\cite{walker2018communicating}, risk~\cite{makhataeva2019safety}, or even robot capabilities~\cite{ostanin2020hri}.
Other work utilizes XR to provide comprehensive visualizations of robot information and the ability to interact with the robot in virtual or augmented spaces~\cite{Hoang_Chan_Lay_Cosgun_Croft_2022}.
Additionally, a number of robot digital twin systems are being developed in XR~\cite{10.1145/3655532.3655538, s24175680}.
Many existing XR robotic systems utilize ROS as their backend for their system's robotic components to take advantage of the various set of tools already developed by other roboticists.
Here, we consider interactive motion planning in reconfigurable environments, allowing users to safely visualize potential paths using a virtual robot.



\subsection{Interactive Motion Planning}

Several prior works have explored the problem of interactive motion planning. Early efforts used only two-dimensional interfaces~\cite{vla-vvaetmp-06} where users could provide regions, or bounding volumes in the environment, to bias motion planning towards or away from these areas while the planner displays its progress to the user and colors the regions based on their perceived usefulness~\cite{dsja-arbsfcrc-14}. 
Lee et al.~\cite{Lee_Lim_Kim_2024} developed a system using Unity and ROS to enable non-expert users to select paths out of a set of available options for robot motion.
Although the motion planning was interactive for users, it did not include any collision detection and the system was not open-source.
Togias et al.~\cite{Togias_Gkournelos_Angelakis_Michalos_Makris_2021} developed a system for remote interactive motion planning for robots used in factories.
In their work, the user was able to modify the paths in VR and check that the planned motions would be satisfactory and without collisions before they were sent to the robot.
Their system, however, does not allow the user to change the configuration of objects in the environment.
Other works~\cite{Karnam_Zelechowski_Cattin_Rauter_Gerig_2025, Chen_Sun_Pollefeys_Blum_2024, Gadre_Rosen_Chien_Phillips_Tellex_Konidaris_2019, quintero2018programming, Rivera-Pinto01092024, ostanin2020hri, ONG2020101820, LOTSARIS2021459} give users control at a lower level, allowing them to manually specify waypoints the robot should visit or plan paths, but again, do not allow environment reconfiguration.
Hernández et al.~\cite{8962156} provide an interactive suite for motion planning allowing users to interact with environment objects in order to specify start and goal constraints.
Users could interact with objects in the environment, however, the set of items was restricted to a set of known objects already setup in the environment. A comparison of system capabilities can be seen in Table \ref{tab:landscape}.

\begin{table*}[ht!]
\centering
\begin{tabular}{lrrrrrr}
\toprule
System & Planner Interface & Trajectory Preview & Environment Modification & ROS 2 Support & Open-Source \\
\midrule
\cite{ONG2020101820} & Preset MoveIt Planner & Yes & Move fiducial marked objects & No & No \\
\cite{LOTSARIS2021459} & Preset MoveIt Planner & Yes & None & No & No \\
\cite{ostanin2020hri} & Preset MoveIt Planner & Yes & None & No & No \\
\cite{Togias_Gkournelos_Angelakis_Michalos_Makris_2021} & Preset MoveIt Planner & Yes & None & No & No \\
\cite{8962156} & User specifies goals & Yes & Move existing task objects & No & No \\
ERUPT & Selectable by user & Yes & Create/modify collision objects & Yes & Yes\\
\bottomrule
\end{tabular}
\caption{Feature comparison of other immersive interactive planning systems.}
\label{tab:landscape}
\end{table*}

While there are many interactive planning systems, none that we could find support ROS2.
Additionally, very few systems support interacting with objects in the environment.
To the best of our knowledge, none of the systems support the creation and modification of virtual objects to evaluate different environment configurations for planning without having to stop the system and move the objects, which 
may interrupt the feeling of being present and having agency in the interactive planning environment~\cite{Skarbez_Brooks_Whitton_2017,Slater_Banakou_Beacco_Gallego_Macia-Varela_Oliva_2022}.
To this end, we propose ERUPT, an interactive extended reality toolkit for motion planning.
ERUPT allows users to interact with the planning environment, including the robot and environment objects, and create motion planning requests without breaking presence.

\section{System Overview}
\label{sec:system_overview}
A system overview is shown in Fig.~\ref{fig:system-diagram}, which we elaborate on below.
Our system consists of two main components, the immersive XR interface and a set of ROS nodes, running on an external computer, that perform planning operations and maintain an environment representation. In this section, we describe the implementation of both of those components and their integration.

\subsection{XR Interface}
The XR interface features a broad range of interaction capabilities with the environment, robot, and motion planner to enable operators to easily reconfigure the environment and plan and visualize robot motions. We implement the XR interface using the Unity game engine and OpenXR (https://www.khronos.org/openxr/), which supports a broad range of modern commercial HMDs. In this work, we use a Meta Quest 3 HMD with its provided controllers to interact with the robot and the environment.

\begin{figure}
    \centering
    \hfill
    \subfigure[Wrist Menu]{
        \includegraphics[height=0.2\textwidth]{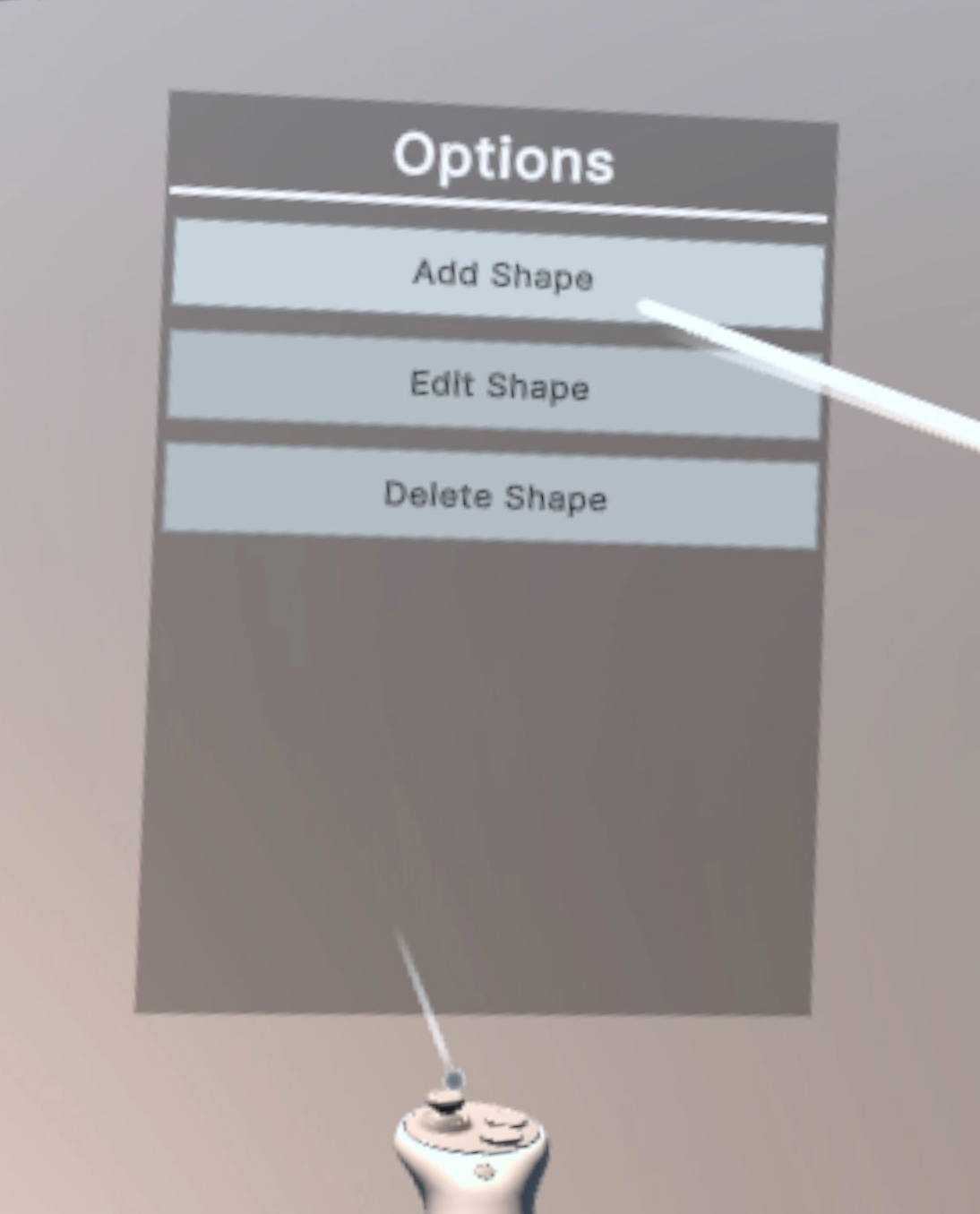}
        \label{fig:wrist-menu}
    }
    \hfill
    \subfigure[Object Selection]{
        \includegraphics[height=0.2\textwidth]{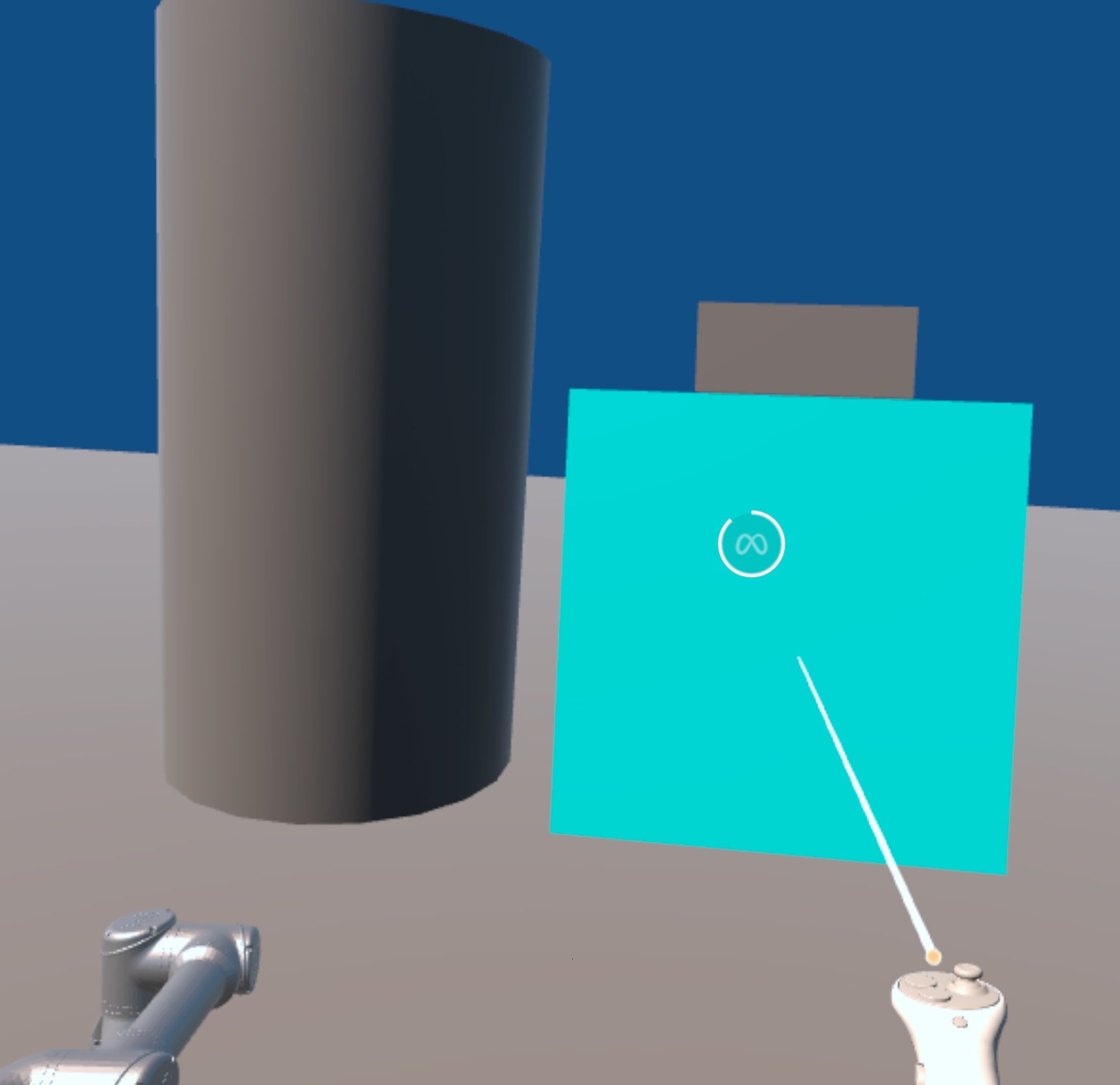}
        \label{fig:shape-selection}
    }
    \hfill
    \hfill
    
    \caption{Through the wrist menu (a), the user can add primitive shapes and edit the geometry of existing shapes. (b) When the user selects an object to edit, it is highlighted in a different material.}
    \label{fig:wrist-menu-and-selection}
\end{figure}

\begin{figure}
    \centering
    \subfigure[AR View]{
        \includegraphics[width=0.4\textwidth]{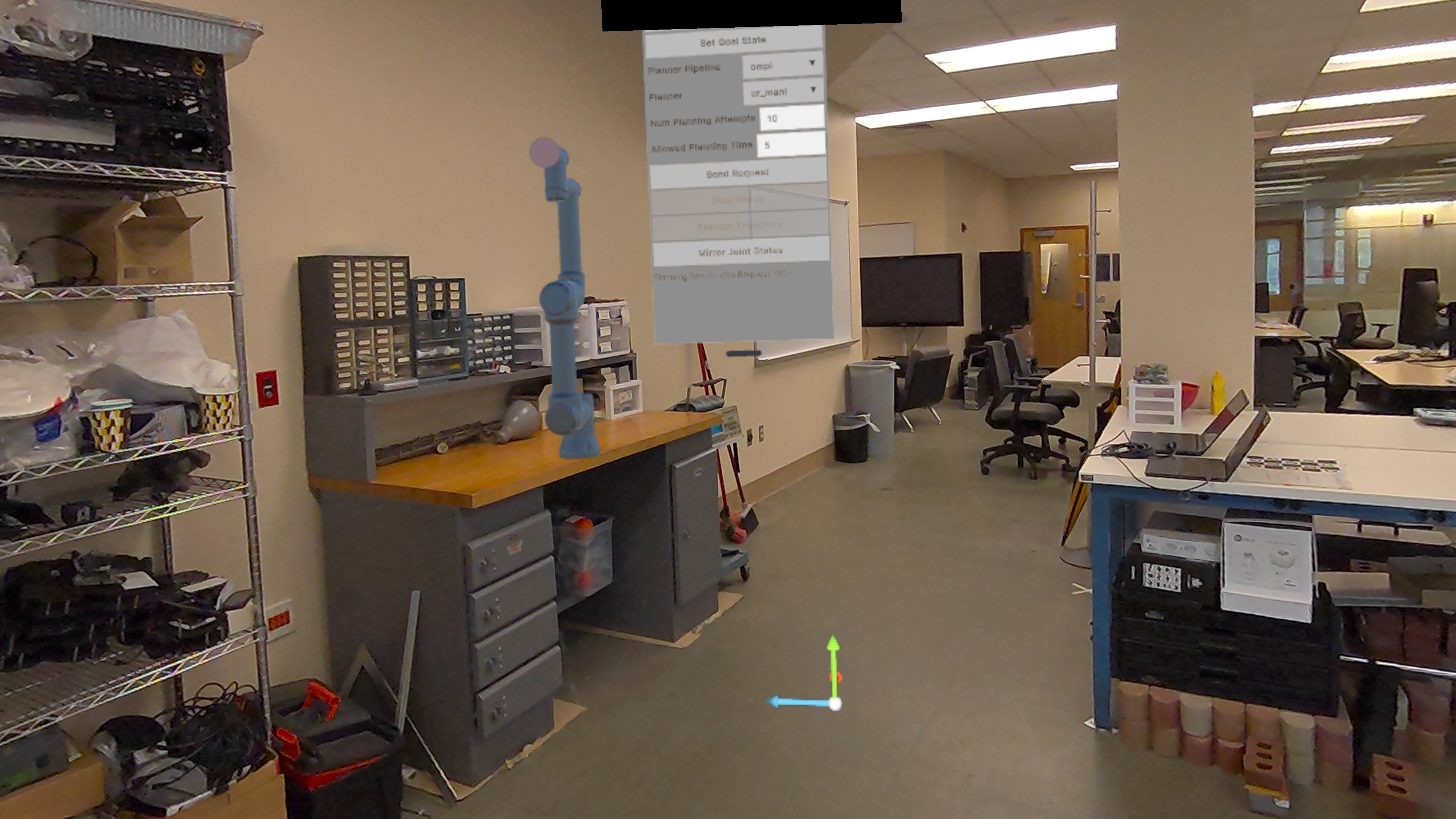}
    }
    \subfigure[MoveIt Planning Scene]{
        \includegraphics[width=0.4\textwidth]{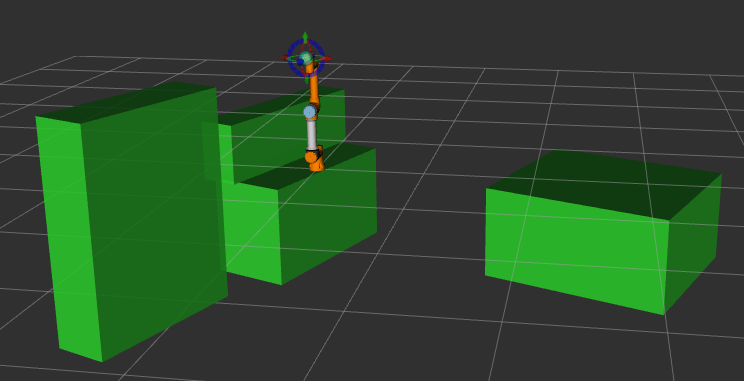}    
    }
    \caption{Environment information can be accessed from the HMD and added to the planning scene. In this example, bounding boxes for furniture are accessed from the Meta Quest 3's environment scan.}
    \label{fig:scene_scan}
\end{figure}

\subsubsection{Interactions with the Environment}
\label{sec:env-interaction}
To configure or reconfigure the environment, each object must be added to the planning scene. The user may have some initial configuration stored on the ROS side. In AR, users also have the option to access information from their HMD's environment representation to expedite environment configuration. For example, the Meta Quest 3 allows users to scan their environment and creates bounding boxes for furniture in the scene. We provide an option to load these into the planning scene (Fig.~\ref{fig:scene_scan}).
Users may alternatively add each object in the XR application.
Each object either initially included in or added to the planning scene in MoveIt is sent over a ROS topic to be added to the virtual environment in XR. To add an object to the planning scene in XR, we provide a toggleable menu attached to the left controller (the \textit{wrist menu}; Fig.~\ref{fig:wrist-menu}) that allows users to spawn primitive shapes. 

Once objects are present in XR, the user may modify them through scaling, rotation, and translation operations. We support two interaction modalities for scaling. For uniform scaling, the user may grab the object with both hands and drag to resize. To scale along a specific axis, for example, and create a rectangular wall from a cube primitive, the user may select the object by pointing the right controller at it and pressing the trigger button. The object will then be highlighted in a different color as an indication to the user (Fig.~\ref{fig:shape-selection}). The user can then open the wrist menu and adjust the scaling along each axis. To adjust the rotation and translation of an object, the user can grab the object and rotate and translate the controller. Deleting an object can be done by selecting it and using the wrist menu.

To ensure consistency between the environment displayed in XR and the planning scene maintained by MoveIt, we attach a script to each object in the environment that listens for changes in scale, rotation, or translation. When a change occurs, the new values are published to ROS.

\begin{figure}
    \centering
    \subfigure[Fiducial Marker]{
        \includegraphics[width=0.35\textwidth]{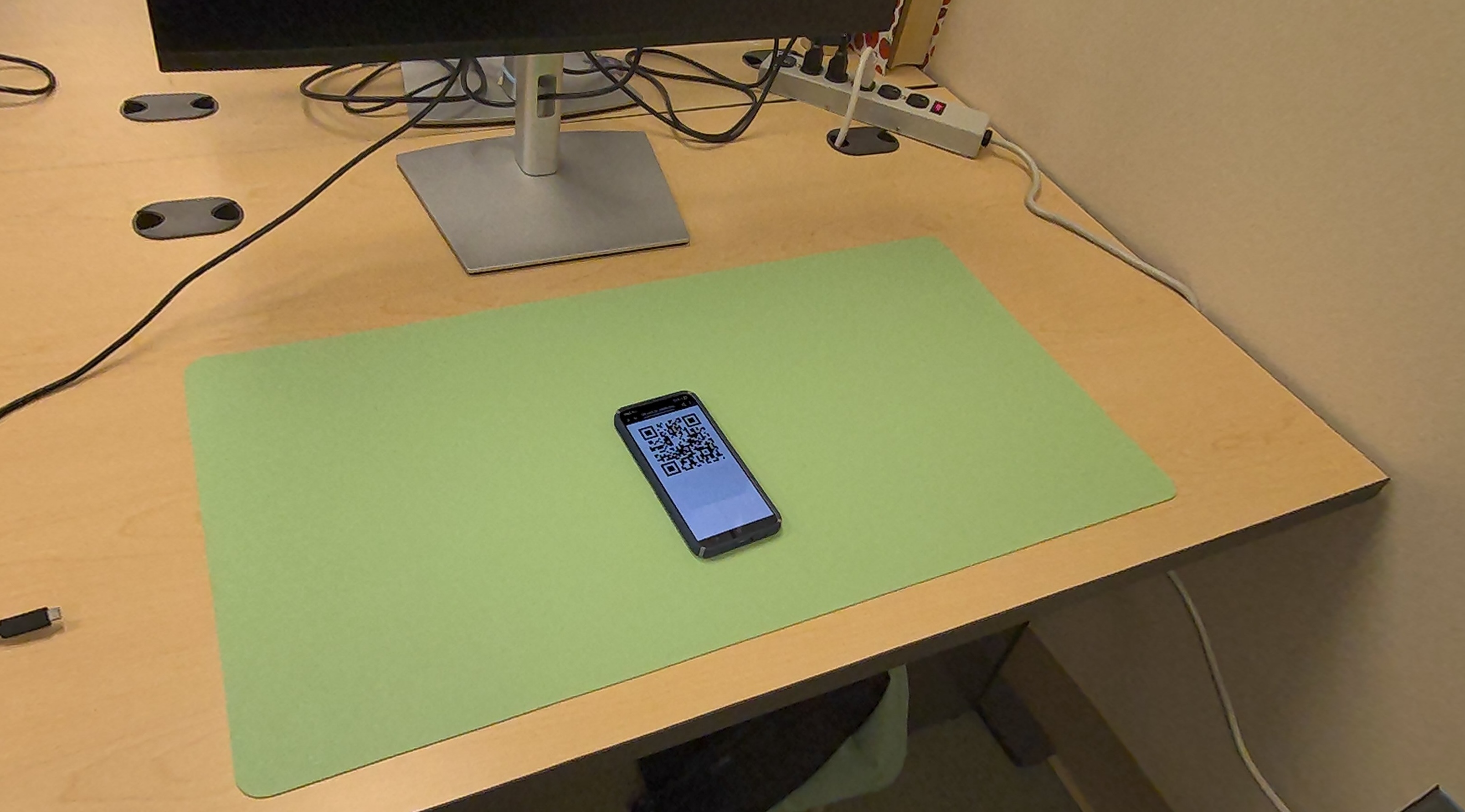}
        \label{fig:qr-code}
    }
    \subfigure[Robot Placement]{
        \includegraphics[width=0.35\textwidth]{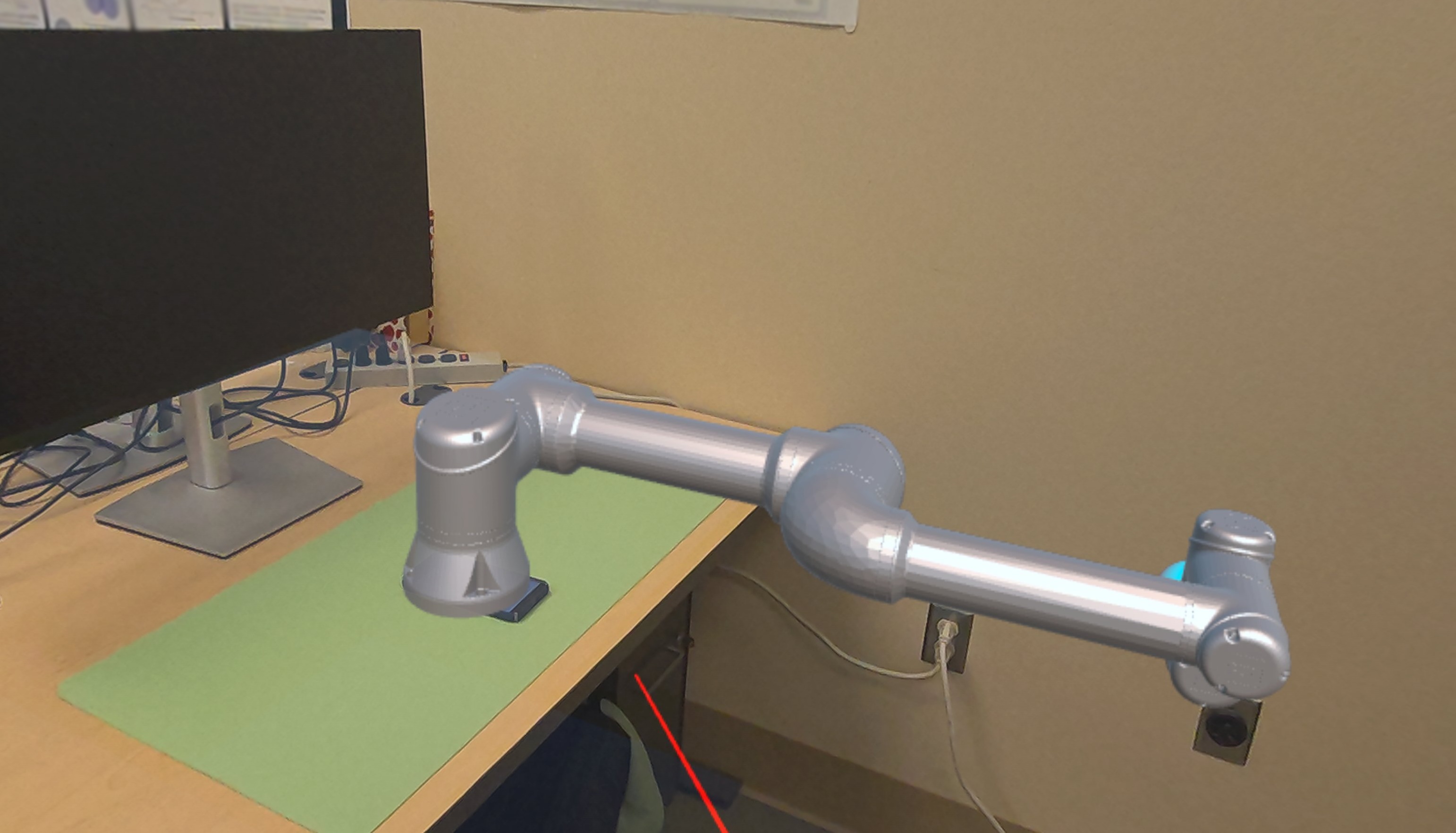}
        \label{fig:robot-placement}
    }
    \caption{An example of placing a robot in an augmented reality environment using a fiducial marker.}
    \label{fig:qr-robot}
\end{figure}

\subsubsection{Interactions with the Robot}
Through the Unity URDF importer (\url{https://github.com/Unity-Technologies/URDF-Importer}), the user can import a robot specified with a URDF file, a common file format to describe the kinematic structure and geometry of a robot. In VR, the user should place the robot at their desired location within the virtual environment. In AR, we provide a method to allow the user to place the robot within the physical environment. This requires access to the scene geometry scanned by the HMD in the environment (for example, the ground plane or tabletops). Some HMDs continuously scan the physical environment to construct these planes while others require the environment to be fully scanned in advance of running the application. The user may place a fiducial marker in the physical environment on top of a scanned plane and the application will detect this location and place the robot's base there.

\begin{figure}
    \centering
    \subfigure[]{
        \includegraphics[width=0.3\textwidth]{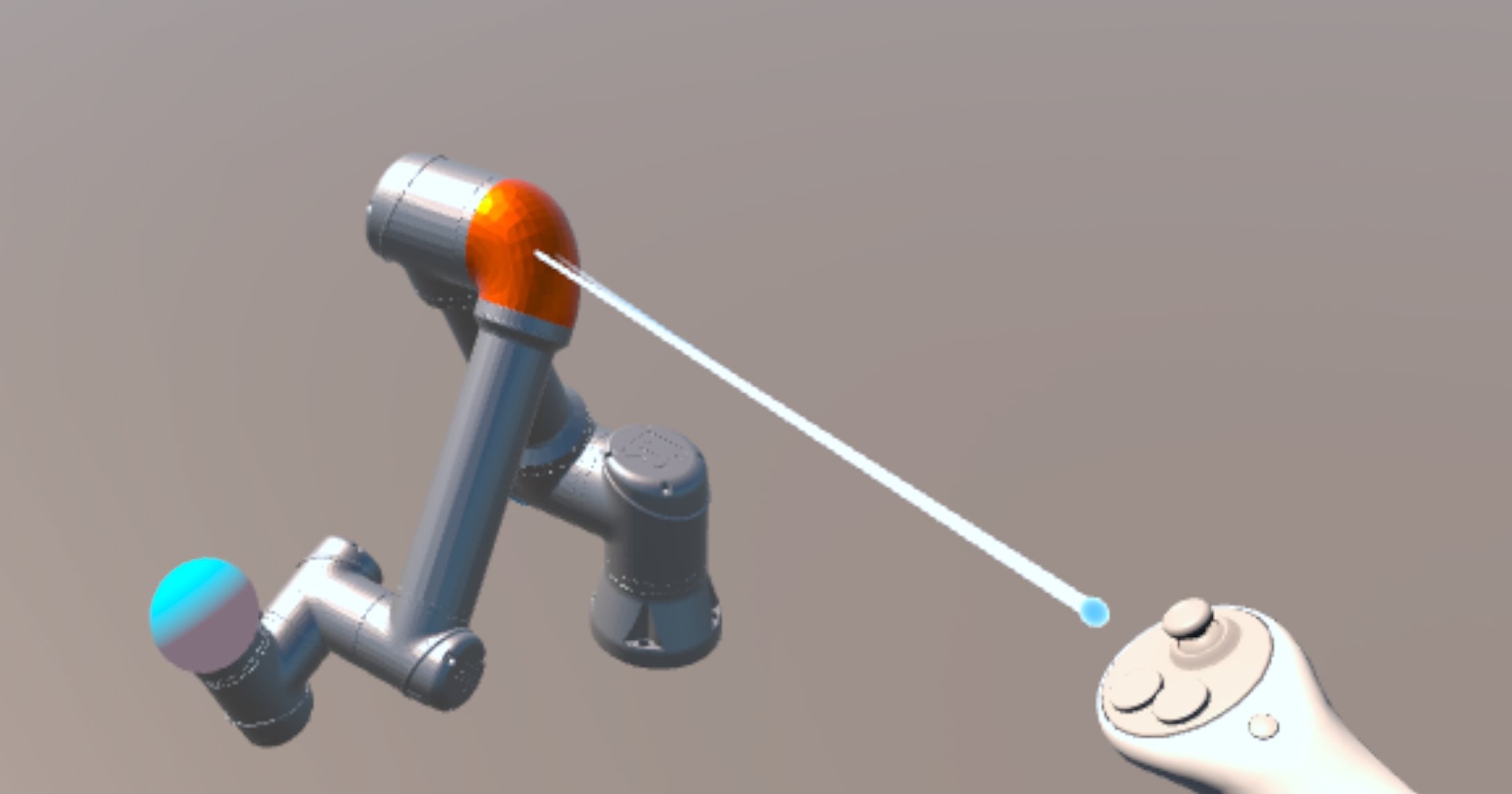}
    }

    \subfigure[]{
        \includegraphics[width=0.3\textwidth]{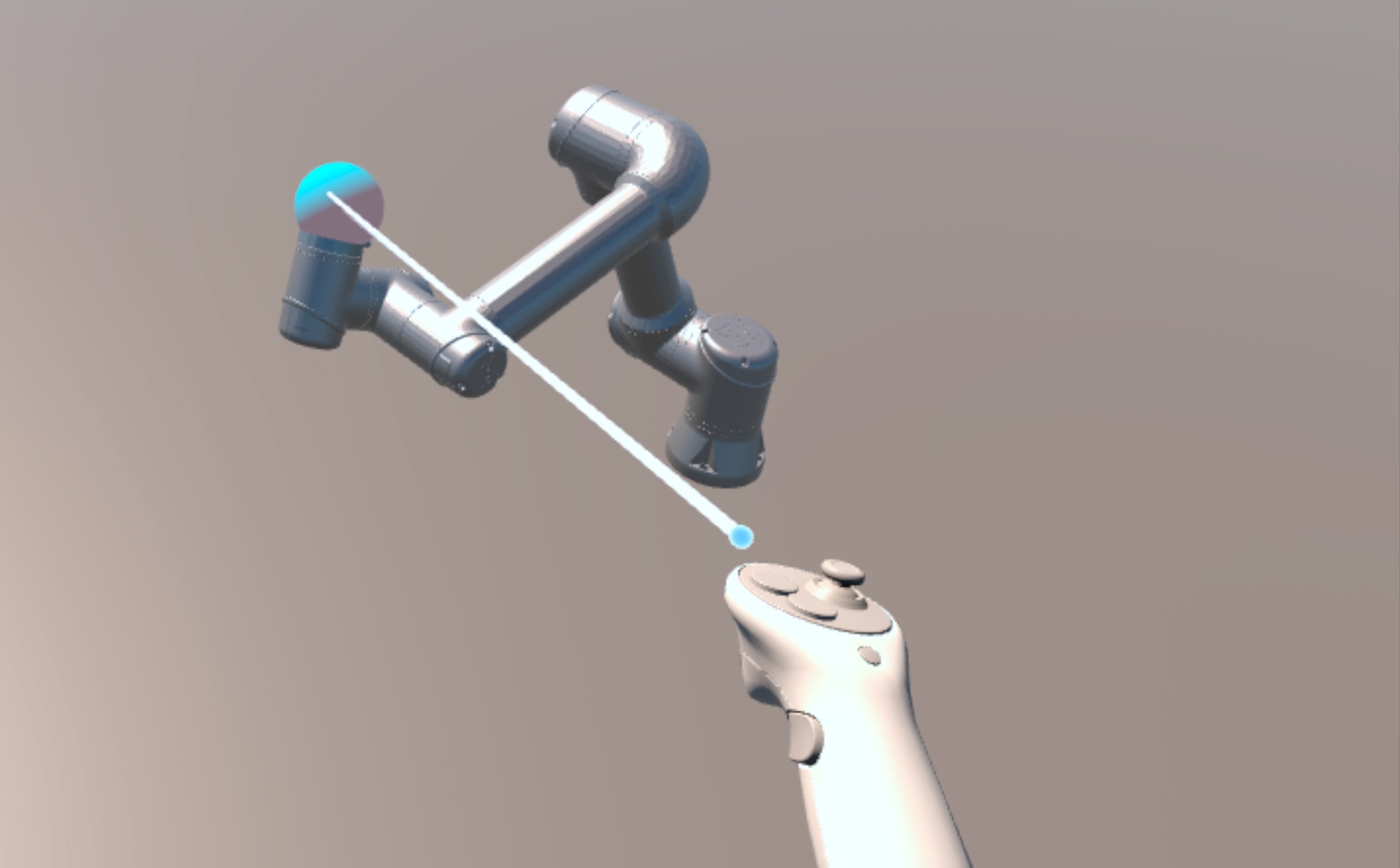}
    }
    \caption{The user may interact with the virtual robot by either (a) directly modifying the joint angles or (b) grabbing and dragging the end-effector using an attached indicator.}
    \label{fig:interaction}
\end{figure}

The user can interact with the robot in two ways (Fig.~\ref{fig:interaction}). First, by directly adjusting the joint angles using a grab and rotate motion with the controller.
Additionally, the user may drag the end-effector of the robot to a desired pose. 
When the end-effector is moved, inverse kinematics is run to compute the satisfying joint positions. All inverse kinematics computations are performed in Unity on the HMD.


\begin{figure}
    \centering
    \includegraphics[width=0.5\linewidth]{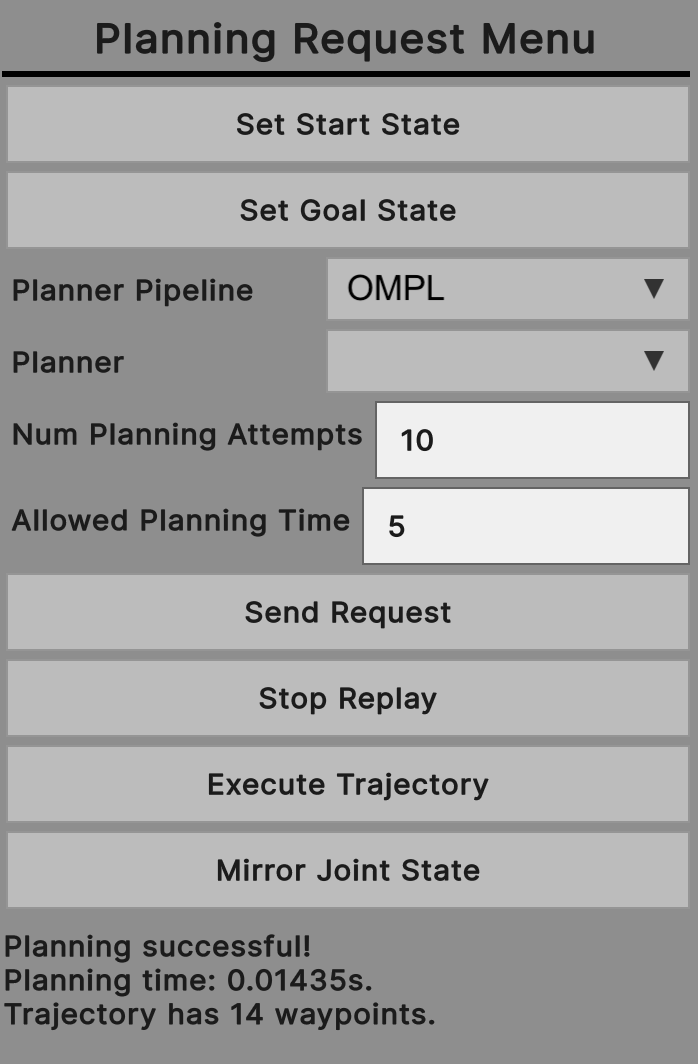}
    \caption{The planning dashboard that allows users to create MoveIt motion planning requests. The user can set the start and goal configurations, the MoveIt planner pipeline and planner, the number of planning attempts, and the maximum planning time. The user can then send the request. The \textit{Stop Replay} button stops the trajectory preview triggered by a successful response. The user may also execute the plan on a physical robot and set the virtual robot to mirror the state of a physical robot.}
    \label{fig:planning-menu}
\end{figure}

\subsubsection{Motion Planning}
Our system features a dashboard (Fig.~\ref{fig:planning-menu}) to allow users to create MoveIt motion planning requests.
The user can set the start and goal configurations by adjusting the virtual robot and pressing the \textit{Set Start State} and \textit{Set Goal State} buttons, respectively. The user can choose from MoveIt planning pipelines and planners. The available options are queried over ROS allowing users to view each planners' behavior. The number of planning attempts and planning time can also be configured. 
After sending the planning request and receiving a response, feedback from the planner is displayed to the user. In the case of success, this includes the planning time and the number of waypoints on the resulting path. In the case of failure, an error code is displayed.

\begin{figure}
    \centering
    \hfill
    \subfigure[]{
        \includegraphics[width=0.18\textwidth]{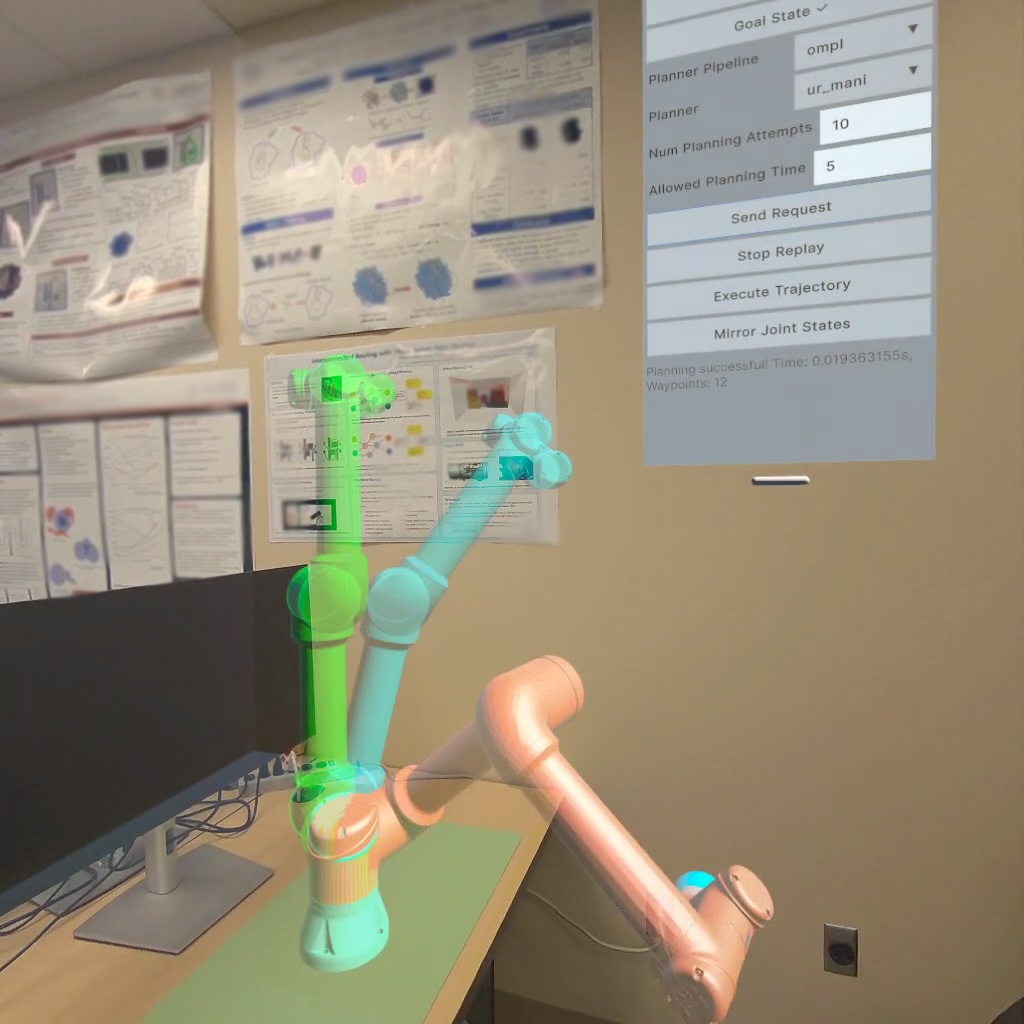}
    }
    \hfill
    \subfigure[]{
        \includegraphics[width=0.18\textwidth]{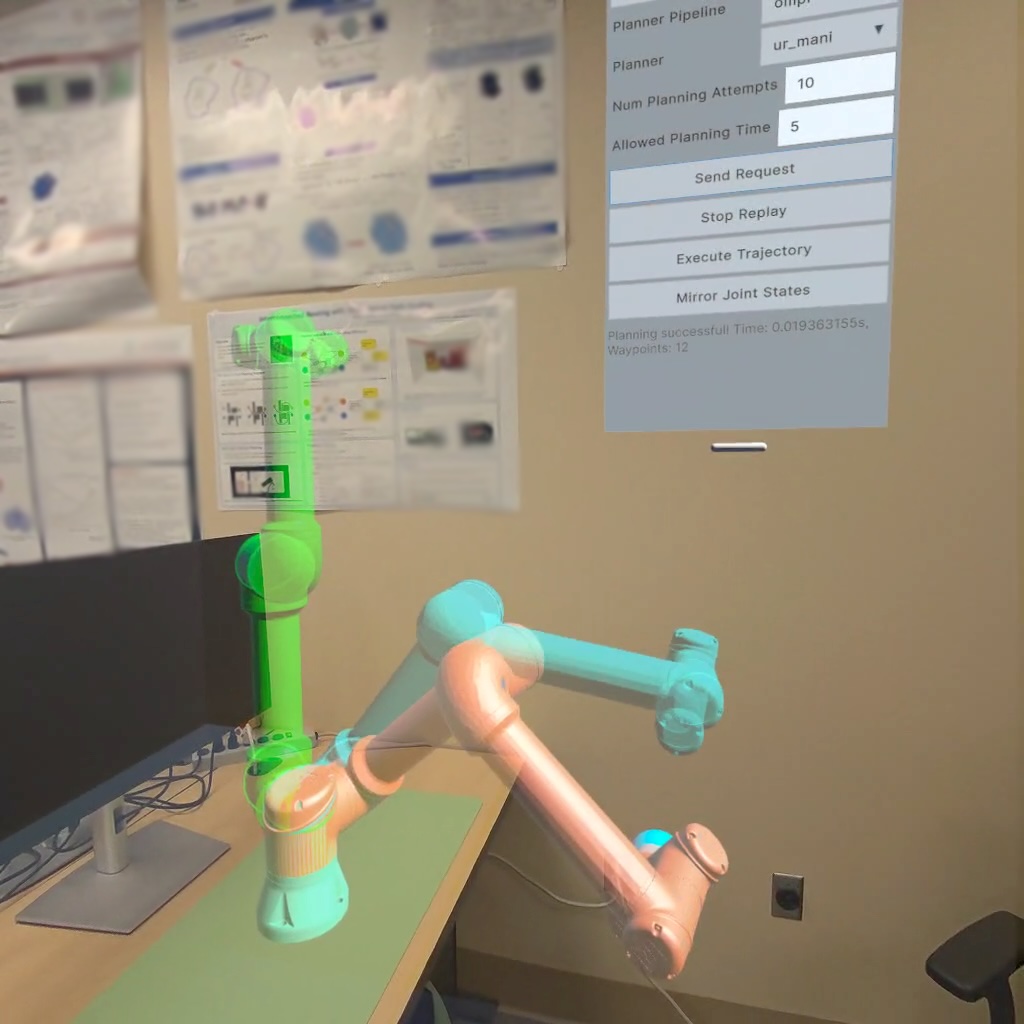}
    }
    \hfill
    \hfill 
    
    \caption{A time-lapse of the result of a motion planning request. The start and goal are shown in green and orange respectively. The cyan robot moves along the output trajectory.}
    \label{fig:preview-timelapse}
\end{figure}

When a successful plan is returned from a request, our system displays a preview (Fig.~\ref{fig:preview-timelapse}) of the virtual robot moving along the path. This allows the operator to evaluate the robot's behavior including, for example, the proximity of the robot to objects in the environment as it moves along its path. To execute a trajectory on a physical robot, the user can press the \textit{Execute Trajectory} button to send the planned trajectory to the robot's controller.

\subsection{ROS Nodes}
Our system relies on a set of ROS nodes which handle the transfer of information between Unity and ROS vital to the system's functionality.
These nodes enable our system to communicate with ROS, plan trajectories for robots, and maintain synchronization between the interactive Unity environment and the planning environment. 
All of the components on the ROS side of our system are implemented in ROS 2 Jazzy (https://docs.ros.org/en/jazzy/index.html).

\subsubsection{Unity - ROS Communication}
To handle communication between Unity and ROS, we utilize the Unity ROS-TCP-Connector, an open source Unity package enabling a Unity application to send and receive ROS messages supporting both ROS and ROS 2.
It functions by creating a TCP socket connection between the Unity program and a ROS node which performs the actual communication in ROS.
This package supports communication through publishing and subscribing as well as services; however, actions are not currently supported.


\subsubsection{MoveIt}
MoveIt, the robotic manipulation platform, provides a tight set of planning integrations with many popular open source planners like the Open Motion Planning Library (OMPL)~\cite{sucan20212ompl}.
Accessing the planning and other functionality of MoveIt can be performed through the use of ROS messages.

\subsubsection{Planning Scene Sync}
At the core of every planning system is the robot and environment representation as it dictates information about valid configurations of the robot as well as information about changes in the environment which may influence planning behavior.
In MoveIt, the Planning Scene is the world representation containing information about the robot state and environment information such as obstacles.

One of the core features of ERUPT is the ability to interact with objects in the environment in XR.
This feature may allow users to feel more present during planning, but also allows rapid evaluation of different environment configurations by providing users the ability to move environment objects around.
In Unity, this capability to move objects in XR is straightforward as described above; however, the MoveIt planning scene must be notified of these changes.
MoveIt provides a \textit{CollisionObject} message which can be published to in order to notify the MoveIt planning scene of added, modified, or removed planning scene objects.

MoveIt provides functionality to get the current planning scene, yet, as more objects are added to the scene, the message can become large and slow to send.
We create a ROS node which manages the communication of planning scene objects and their updates on the MoveIt side, and communicates the updates with Unity.
The node keeps track of the objects which have been communicated to Unity and checks if any objects have been modified on the MoveIt side.
If modifications are detected, the node communicates to Unity only the modified objects ensuring no redundant messages are sent, removing the burden on the HMD to process the potentially large planning scene updates.

As discussed earlier in Section~\ref{sec:env-interaction}, we attach a script to each interactable object which efficiently notifies the planning scene of any changes to them.
Whenever new objects are added by the user in XR or in MoveIt, the message notifies the planning scene that it should add a new collision object with the specific parameters of the new object.
When objects are resized, there is unfortunately no efficient way to scale the objects provided by MoveIt, so we must notify the planning scene to delete the object and add the resized version.
If the pose, translation, or rotation, of the object is changed, we can notify the planning scene to only update the pose of the object with the new pose.
Finally, if the object is deleted, the planning scene is notified to remove the object.
The use of these efficient operations allows the system to quickly update the MoveIt environment representation while the user is moving objects around.
We evaluate the performance impacts of these operations in terms of rendering performance and communication latency later in Section~\ref{sec:system-eval}.

\begin{figure*}[ht!]
    \centering
    \subfigure[]{
        \includegraphics[height=0.14\textwidth]{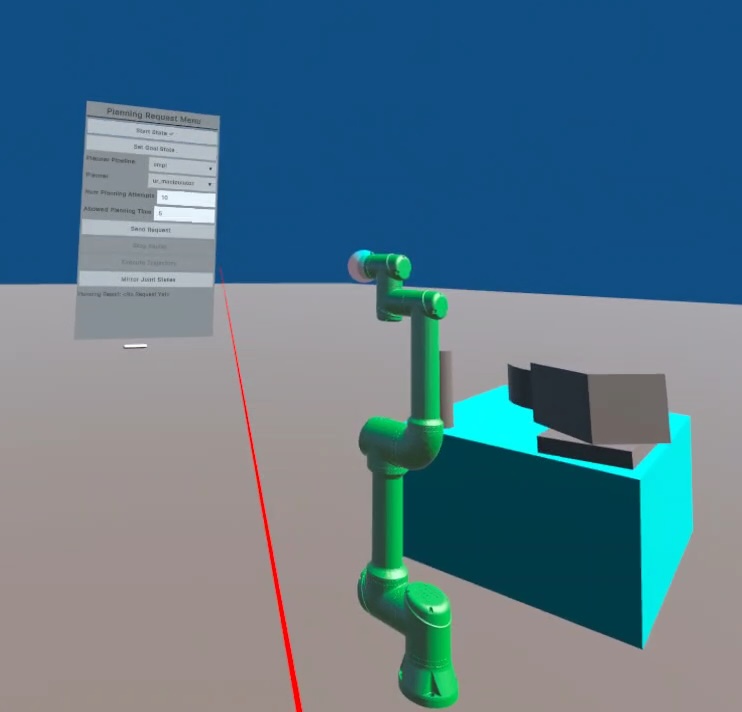}
    }
    \subfigure[]{
        \includegraphics[height=0.14\textwidth]{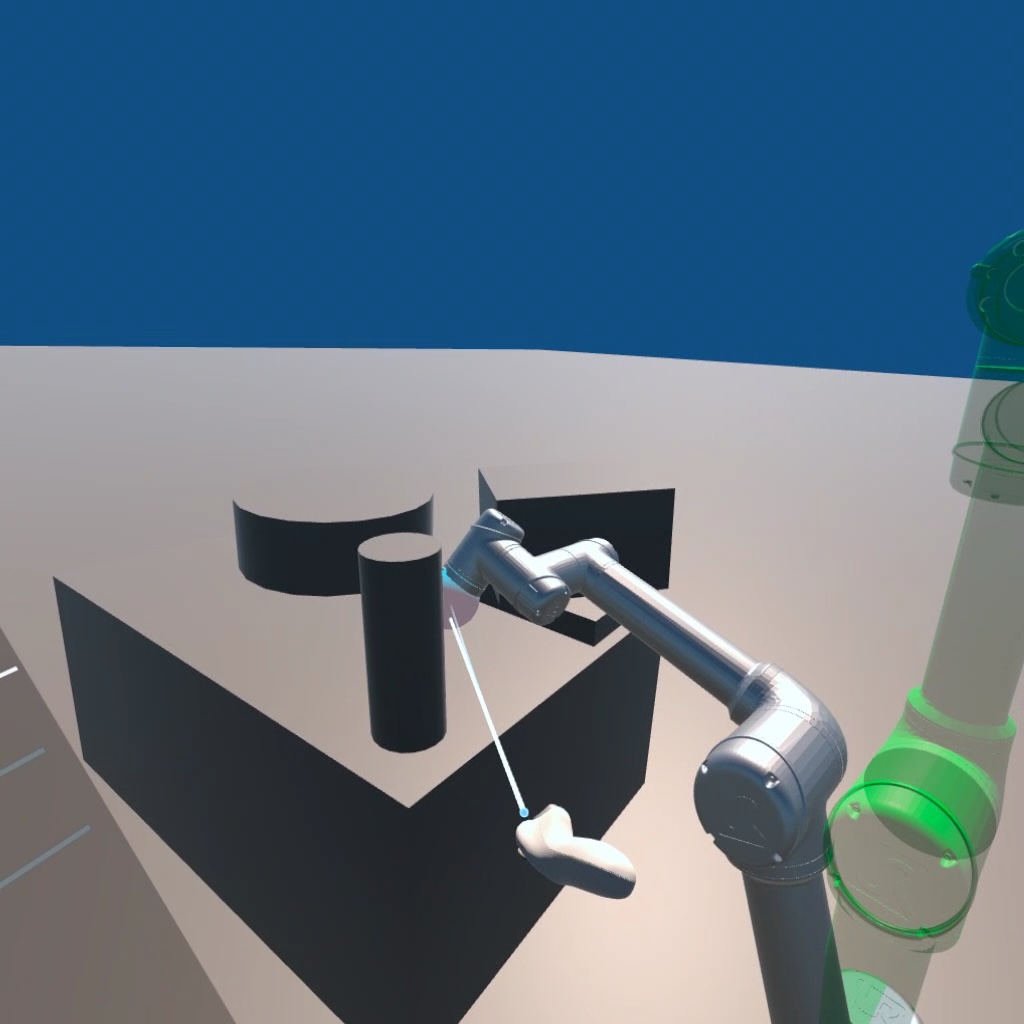}
    }
    \subfigure[]{
        \includegraphics[height=0.14\textwidth]{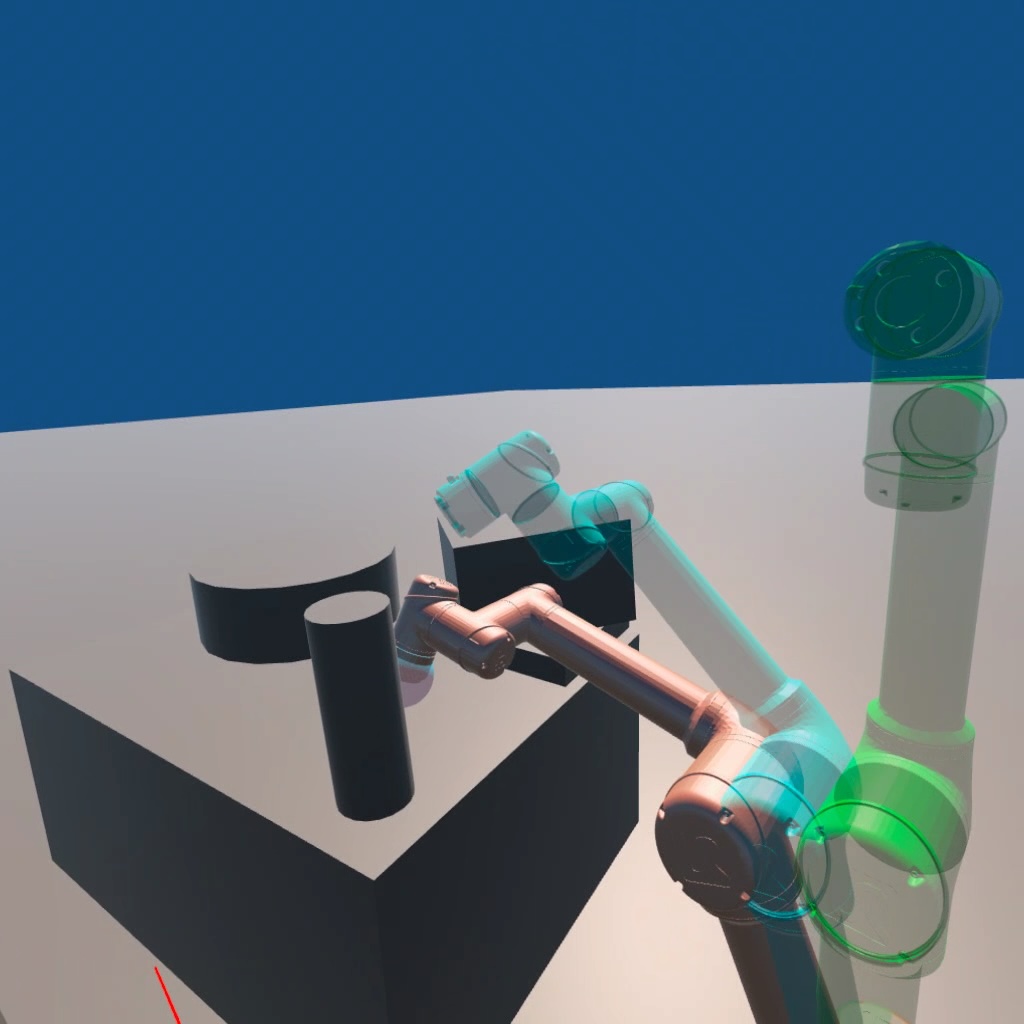}
    }
    \subfigure[]{
        \includegraphics[height=0.14\textwidth]{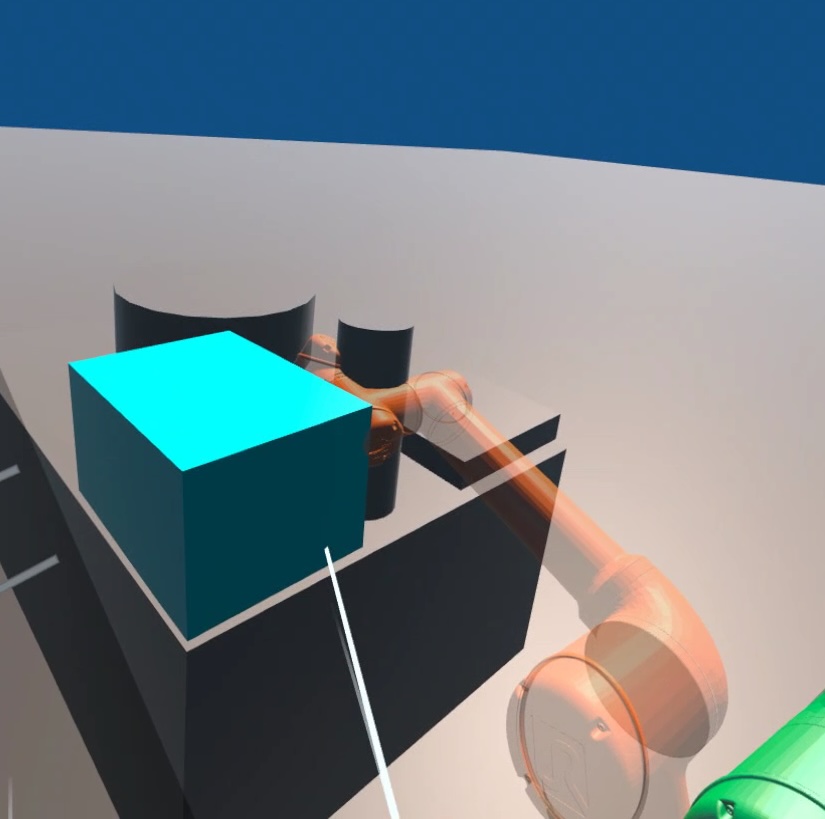}
    }
    \subfigure[]{
        \includegraphics[height=0.14\textwidth]{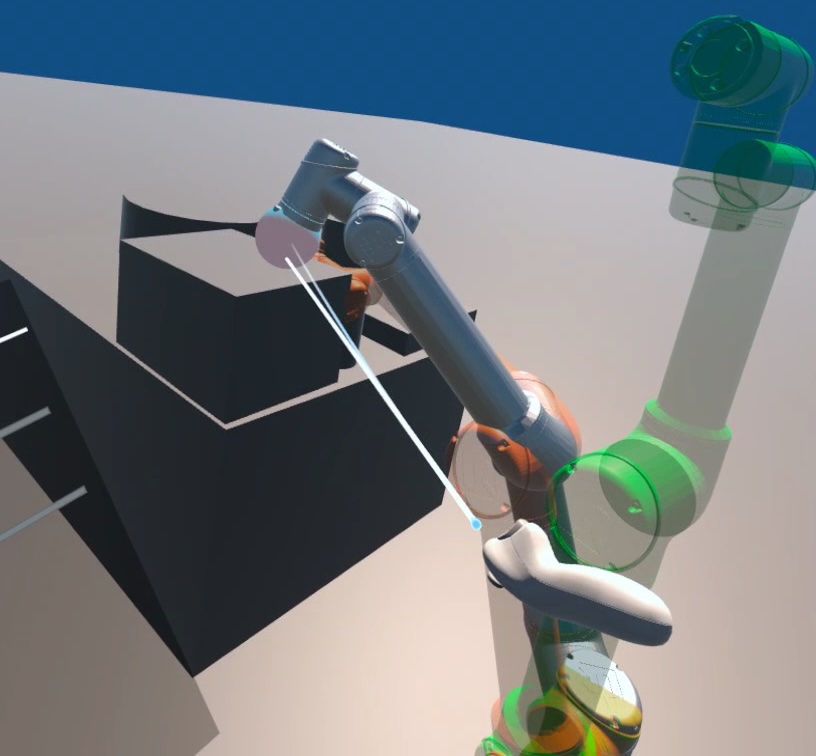}
    }
    \subfigure[]{
        \includegraphics[height=0.14\textwidth]{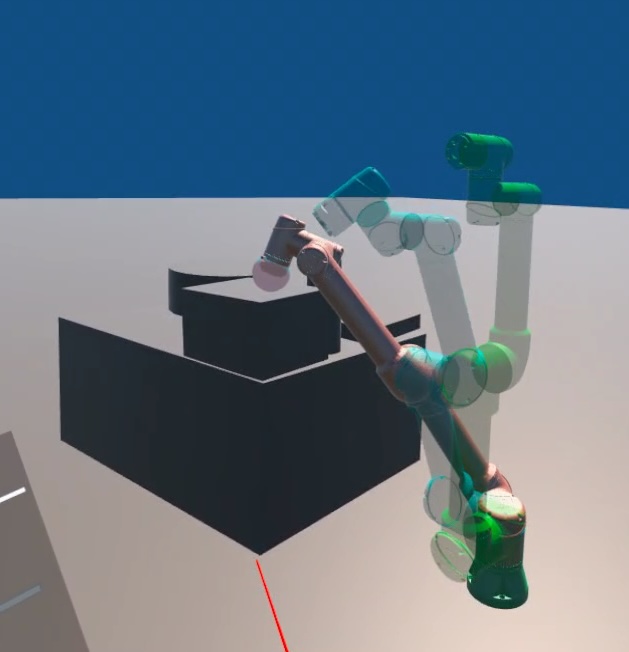}
    }
    \caption{A time-lapse of rearranging a virtual environment to simulate different tasks and replanning. (a) The initial environment setup with multiple items on a table. (b) The user sets the goal pose to simulate reaching an empty spot on the table (e.g., to place an object). (c) The output trajectory is shown in cyan. (d) The user reconfigures the environment objects to simulate rearranging a workspace. (e) The user selects a new goal position to simulate reaching the top of a box. (f) The new output trajectory is shown in cyan.}
    \label{fig:rearrangement}
\end{figure*}

\begin{figure*}[ht!]
    \centering
    \subfigure[]{
        \includegraphics[height=0.14\textwidth]{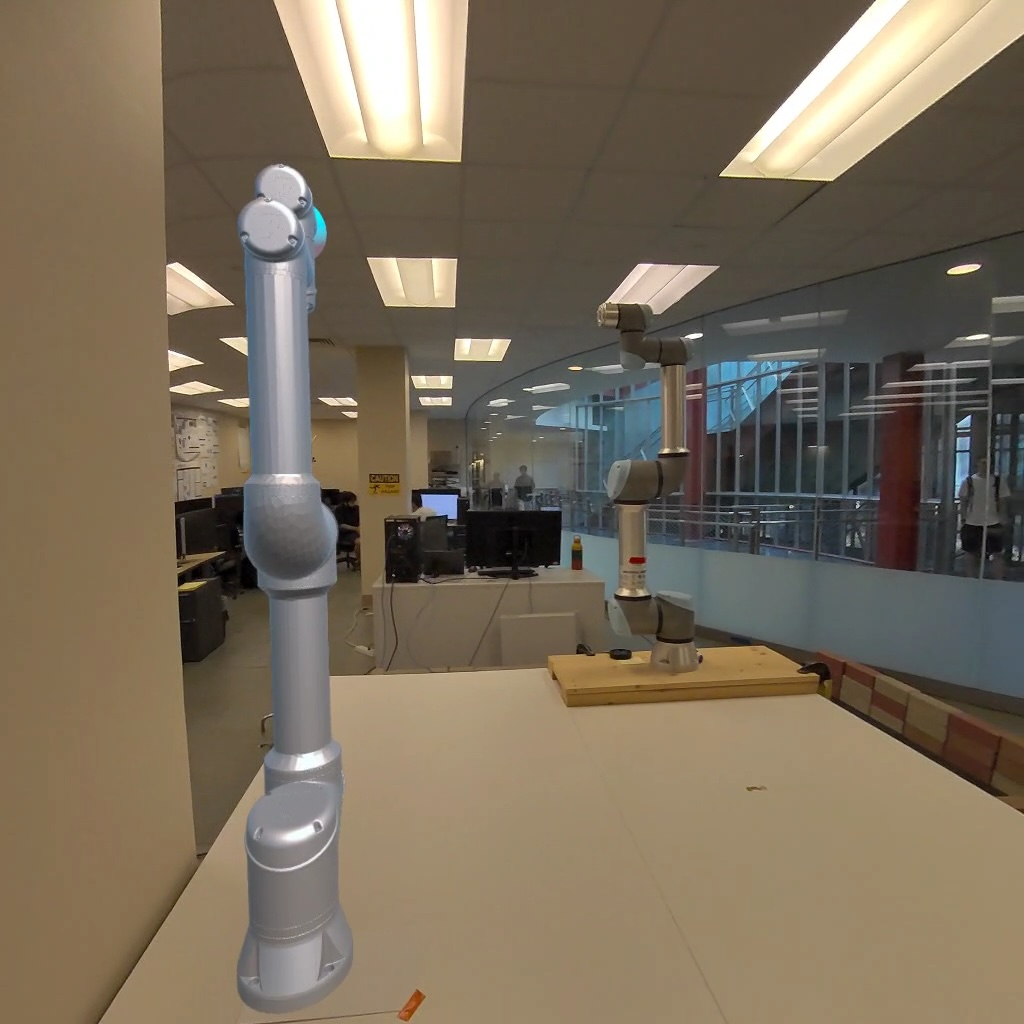}
    }
    \subfigure[]{
        \includegraphics[height=0.14\textwidth]{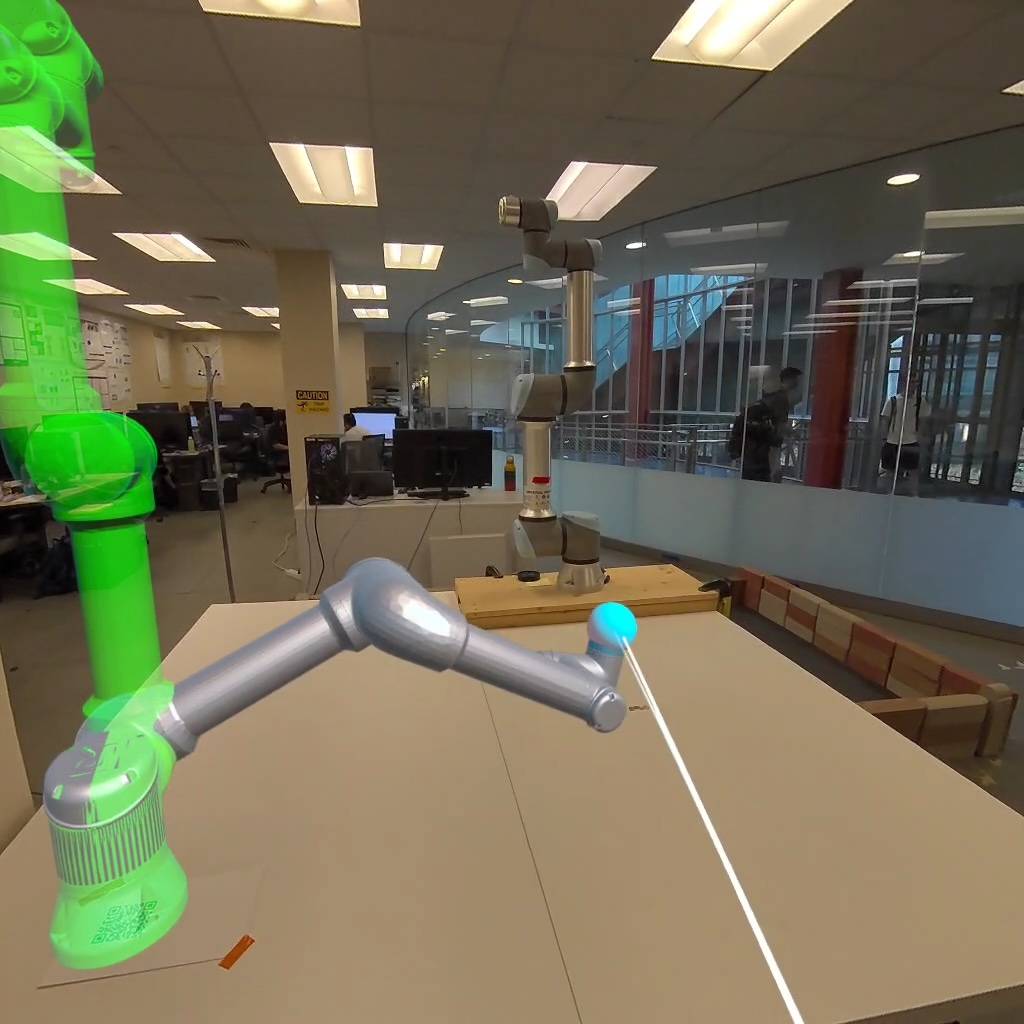}
    }
    \subfigure[]{
        \includegraphics[height=0.14\textwidth]{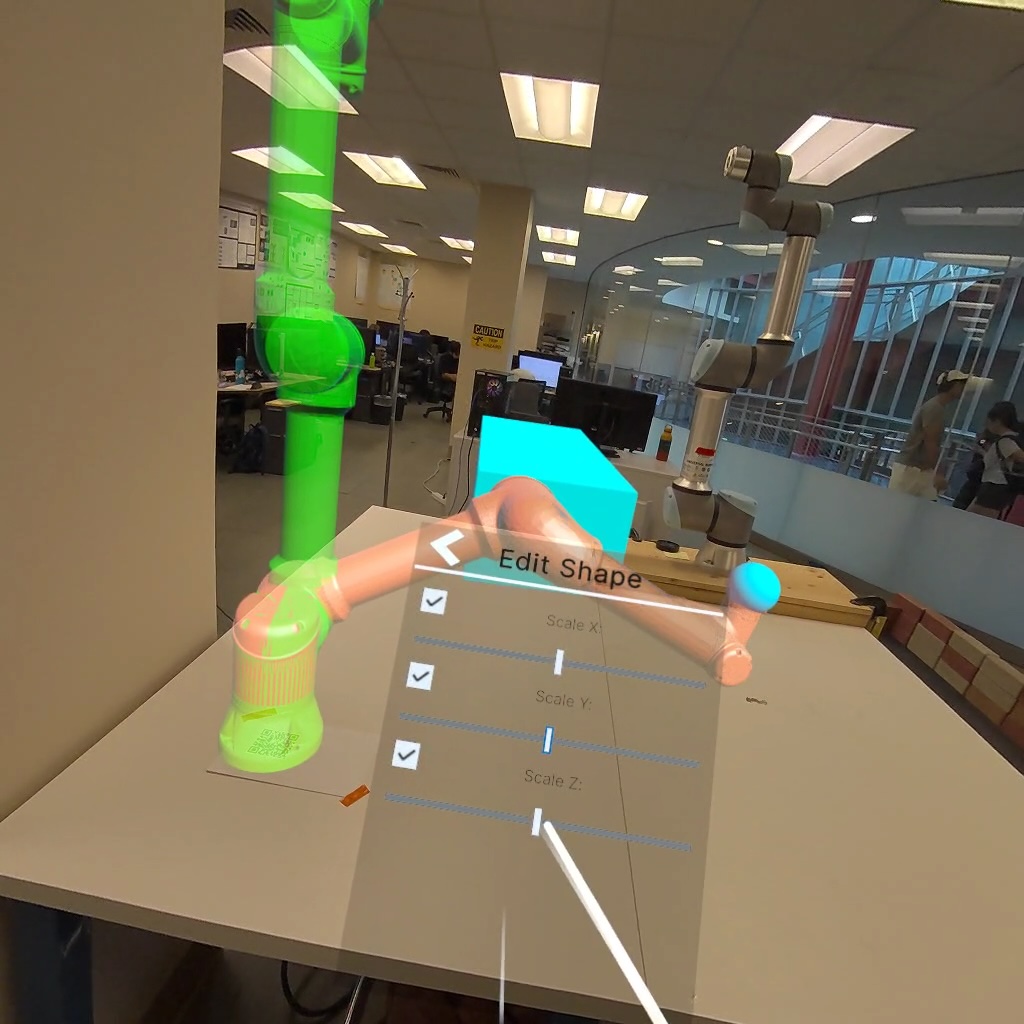}
    }
    \subfigure[]{
        \includegraphics[height=0.14\textwidth]{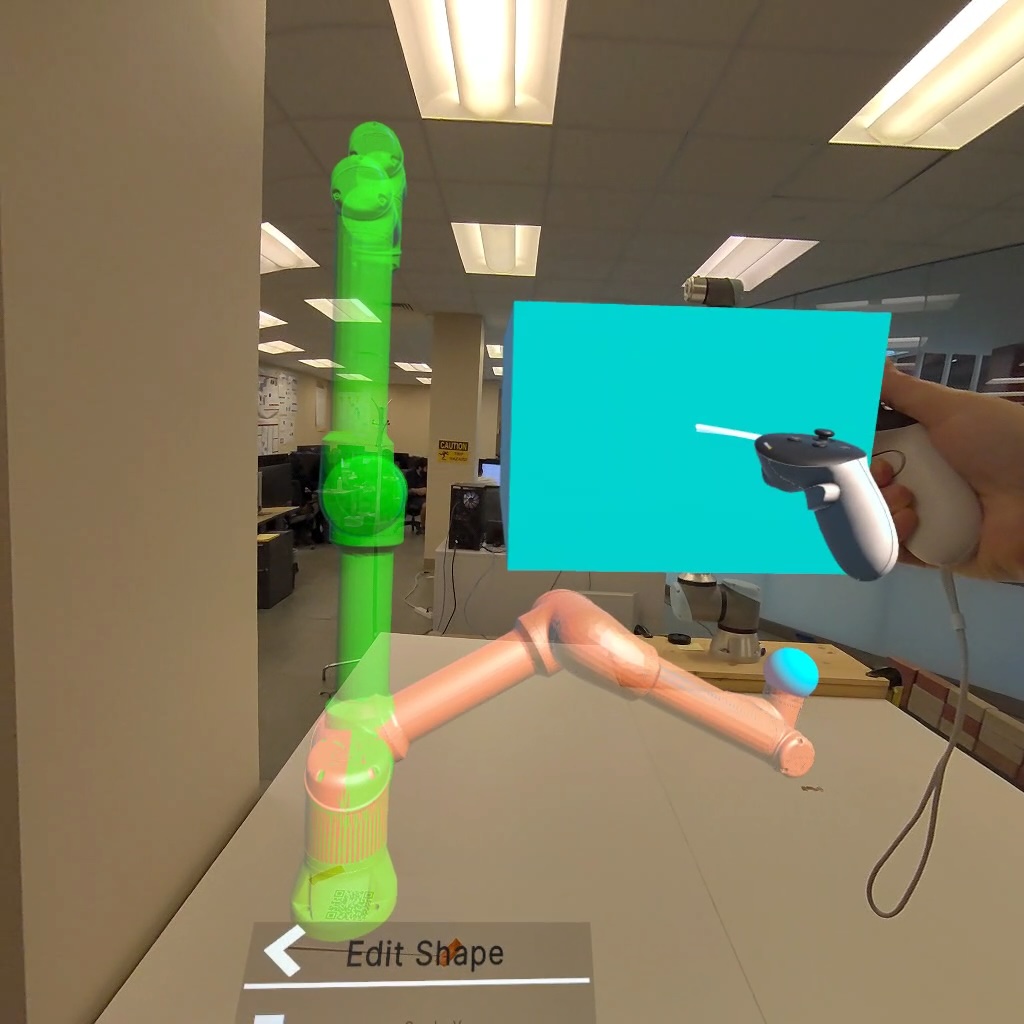}
    }
    \subfigure[]{
        \includegraphics[height=0.14\textwidth]{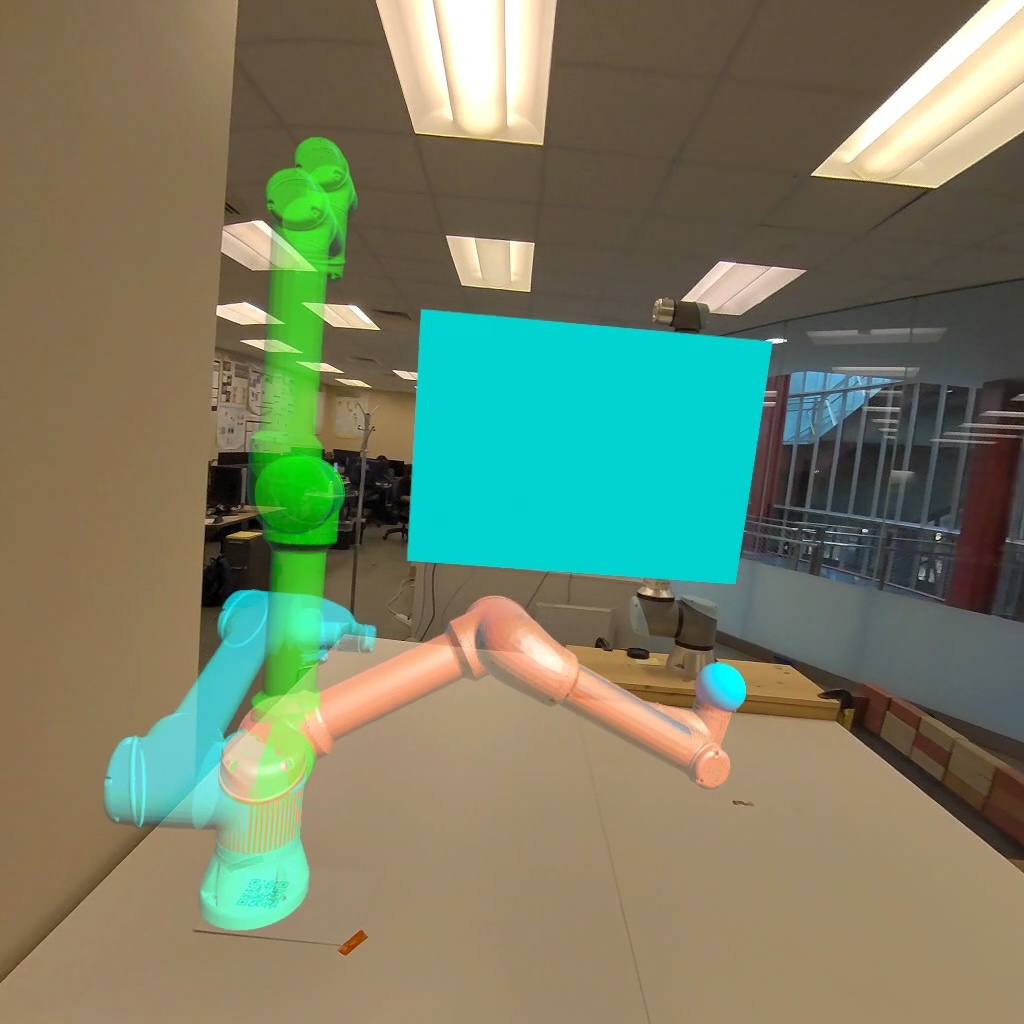}
    }
    \subfigure[]{
        \includegraphics[height=0.14\textwidth]{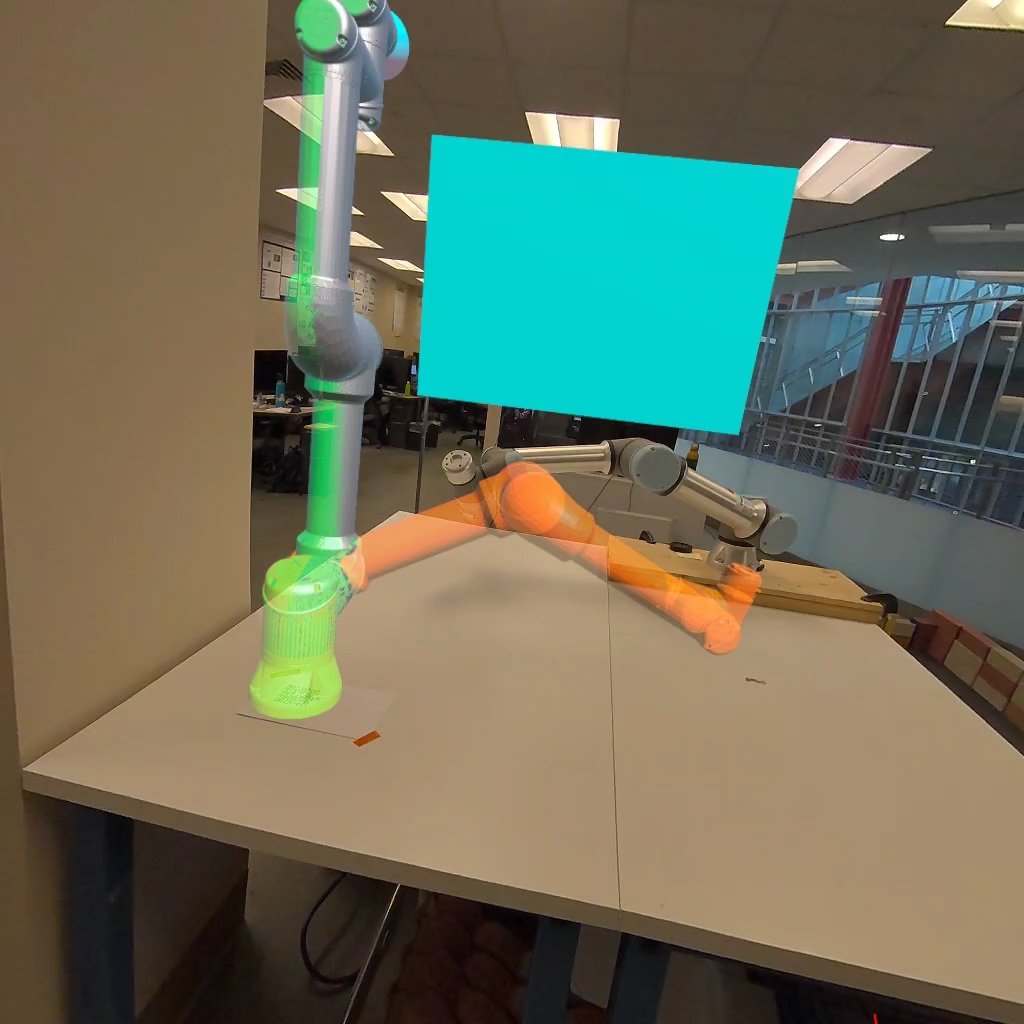}
    }
    \caption{A time-lapse of previewing a trajectory before executing on a physical robot. (a) The user places the robot on the table, and (b) sets the start and goal configurations. (c) The user creates a virtual obstacle and (d) places it where they desire. (e) After sending a planning request, the cyan robot shows the planned trajectory. (f) The user executes a satisfactory trajectory on the physical robot.}
    \label{fig:phys-demo}
\end{figure*}

\subsubsection{Motion Planning}
Motion planning is performed through a \textit{MotionPlanRequest} message where the user can specify information about the planning problem such as the starting state, goal constraints, as well as which planning algorithm to use.
The response to the planning request contains information about any errors that occurred during planning, as well as the trajectory if planning was successful.
The successful trajectories can then be executed through MoveIt or can be directly sent to a robot controller. 

\begin{table*}[t]
  \centering
  \caption{Planning scene update latency and frame rate by operation and object type. FT denotes frame time.}
  \label{tab:latency-fps}
  \begin{tabular}{llrrrrrrrrr}
    \toprule
    & & \multicolumn{3}{c}{Latency (ms)} & \multicolumn{6}{c}{FPS} \\
    \cmidrule(lr){3-5} \cmidrule(lr){6-11}
    Operation & Object & Mean & Median & Std.\ Dev. & Mean & Median & Min & 1\% Low & Std.\ Dev. & Mean FT (ms) \\
    \midrule
    \multirow{4}{*}{Create} & Cube         &   38.2 &   32.0 &  31.7 & 121.6 & 119.0 & 59.0 & 70.3 & 18.3 &  8.40 \\
                            & Mesh (512)   &   62.0 &   58.0 &  19.3 & 122.1 & 118.8 & 52.7 & 60.5 & 22.6 &  8.45 \\
                            & Mesh (5000)  &  318.6 &  314.3 &  71.2 & 121.8 & 119.5 &  9.7 & 11.1 & 30.6 &  9.90 \\
                            & Mesh (20000) & 1734.6 & 1579.7 & 300.0 &  99.7 & 101.5 &  1.0 &  1.0 & 43.3 & 67.30 \\
    \midrule
    \multirow{4}{*}{Move}   & Cube         &   40.9 &   33.6 &  23.5 & 121.4 & 119.2 & 54.1 & 69.5 & 17.1 &  8.40 \\
                            & Mesh (512)   &   39.9 &   32.7 &  24.5 & 121.5 & 119.2 & 49.2 & 69.9 & 17.7 &  8.39 \\
                            & Mesh (5000)  &   39.3 &   32.7 &  23.7 & 121.6 & 119.2 & 41.4 & 68.5 & 18.3 &  8.40 \\
                            & Mesh (20000) &  116.2 &   37.8 & 207.1 & 116.2 & 115.0 & 23.3 & 59.6 & 18.9 &  8.85 \\
    \midrule
    \multirow{4}{*}{Delete} & Cube         &   40.1 &   38.7 &  12.5 & 121.3 & 119.6 & 56.1 & 72.1 & 16.4 &  8.38 \\
                            & Mesh (512)   &   40.0 &   38.5 &  22.3 & 121.5 & 119.6 & 54.2 & 75.9 & 16.8 &  8.38 \\
                            & Mesh (5000)  &   46.1 &   41.0 &  20.2 & 121.5 & 119.7 & 43.0 & 74.1 & 16.1 &  8.38 \\
                            & Mesh (20000) &   43.3 &   39.4 &  31.9 & 115.2 & 114.1 & 23.9 & 64.2 & 18.5 &  8.90 \\
    \midrule
    Request & --- & --- & --- & --- & 110.31 & 113.0 & 95.0 & 95.0 & 8.87 & 9.13 \\
    Baseline   & --- & --- & --- & --- & 120.7 & 119.5 & 49.4 & 68.5 & 14.4 & 8.41 \\
    \bottomrule
  \end{tabular}
\end{table*}

\section{System Evaluation}
\label{sec:system-eval}
We evaluate our system to understand the impact that our system has on rendering performance in terms of frame rate and on latency in communication with ROS and MoveIt to update the planning scene.

\subsection{Rendering Performance}
\label{sec:fps}
To understand the runtime cost of ERUPT's interaction modes on the HMD, we measured the rendering performance across 14 scenarios including an idle baseline, sending motion planning requests, and creating, moving, and deleting collision objects.
The baseline scenario captures the users in an empty scene with no active interaction for 30 seconds.
The planning request scenario looks at the interval of time surrounding 10 motion planning requests to MoveIt, including receiving the trajectory from MoveIt and the preview of the trajectory.
The remaining scenarios target object operations for both primitive shapes and meshes, including adding 50 objects, moving each object 10 times, and then removing $50$ objects.
Three different sphere meshes were used, which varied in tessellation density, yielding 512, 5,000, and 20,000 triangles respectively. No mesh data was cached or deduplicated.
Tests ran on the Meta Quest $3$ at $120$ Hz refresh, with the application built in Unity $6.2$ using OpenXR.
Per-frame timing was logged on the HMD and we report the mean, median, minimum, 1st-percentile low, standard deviation, and mean frame time in Table~\ref{tab:latency-fps}. This data was collected by the authors.

The baseline condition achieves a mean of $120.7$ FPS ($8.41$ ms frame time), effectively holding the HMD's native refresh rate.
Primitive object operations, creating, moving, and deleting cubes, are virtually indistinguishable from the baseline with each maintaining a mean of about $120$ FPS with $1\%$ lows at or above $70$, showing that the operations with primitive objects impose a small cost on rendering.
Mesh operations are the most expensive interactions with the 20,000 triangle mesh creation, incurring the largest sustained cost (mean $99.7$ FPS, $1\%$ low $1.0$) while moving and deleting meshes fall between the primitive cases and mesh creation.
The motion planning request scenario achieves a mean of about $110$ FPS which is the second-lowest.
This may be due to the large message containing the trajectory being processed and prepared for preview.

\subsection{Communication Latency}
We analyze communication latency to measure the time necessary to update the planning scene when creating, deleting, and moving objects in the scene. We measured the round-trip time to send ROS messages from the HMD to the computer managing the planning scene and receive an acknowledgment back to the HMD. We show the results of the same scenarios described in Sec.~\ref{sec:fps} in Table~\ref{tab:latency-fps}.

The latency for adding a cube averages $38.2$ ms.
The latency for adding a mesh depends heavily on its size. The creation time latency for a 20,000 triangle mesh averaged 1,734.6 ms with a standard deviation of 300.0 ms. After creation, meshes can be moved or deleted with relatively low latency since only their object identifier and transform, in the case of movement, need to be sent.
While the latency for adding a mesh is higher due to the higher complexity of the shape, which requires more data to be sent, the planning scene is unlikely to be stale when a planning request is received.
We note that some variance can be attributed to network jitter as the mean RFC 3550 jitter was 21.6 ms over the span of testing.

\section{Demonstrations}
\label{sec:demonstrations}
We provide a set of demonstrations of the ERUPT system which highlight a set of core features for interacting with robots in extended reality. These demonstrations show the ability of our system to be used to assess different environment configurations before deploying a robot and previewing trajectories for execution on physical robots.

\subsection{Environment Reconfiguration}
A key benefit of using an XR system to visualize robot behavior is the ability to easily reconfigure an environment and quickly evaluate of robot motion plans. Using an XR interface allows users to naturally grab and drag objects and place them where desired. This is useful in scenarios where a human is rearranging objects in a shared human-robot workspace or testing different placements of objects in a factory to evaluate a new floor layout for efficiency. Using ERUPT, the human can reprogram the robot's behavior to account for the new workspace configuration.

Fig.~\ref{fig:rearrangement} shows an example of rearranging a virtual environment and replanning a robot's trajectory. The user plans an initial path for the robot to reach an empty space on a table, then rearranges the environment objects, and plans a new path for the robot to reach the top of a box on the table that was moved. After each path is successfully planned, it is displayed to the user.

\subsection{Trajectory Previewing}
Previewing a robot's planned trajectory in XR allows the user to view the robot's motion in three dimensions.
The user can move around the environment while previewing the trajectory to evaluate the robot's proximity to obstacles in the environment. Additionally, placing virtual obstacles can be used to affect the robot's behavior and convey the user's preferences. For example, the user may place a virtual obstacle in a \textit{no-go zone} where a physical obstacle will eventually be placed or where a human is likely to be present.

We show an example of this in Fig.~\ref{fig:phys-demo}. The user first places the virtual robot on a table in augmented reality using the fiducial marker placement method described in Sec.~\ref{sec:system_overview}. In this scenario, the user places the robot on the same table as a physical robot of the same type. The user then creates an obstacle in the environment and scales it to the desired shape. The start configuration is set to the physical robot's current configuration. The user drags the virtual robot by its end-effector to set the desired goal location and sends a planning request. The output trajectory is displayed as a preview to the user with a virtual robot. After inspecting the path and ensuring the robot remains an acceptable distance away from the virtual obstacle, the user executes the trajectory on the physical robot. See supplemental video for examples of these demos.

\section{Limitations}
\label{sec:limits}
Although ERUPT supports importing mesh geometry into the planning scene, our current approach lacks photorealism in XR and struggles to create large meshes at runtime. Moving forward, we plan to explore the ability to support more visually realistic simulation environments and measure the system's ability to scale to large numbers of objects in the scene.

Additionally, our system currently only supports static obstacles.
We would like to extend the planning capabilities supported by our system to dynamic objects so users can view planning behavior in the presence of moving objects.
This should be possible through the use of MoveIt's Hybrid Planner (\url{https://moveit.picknik.ai/main/doc/concepts/hybrid\_planning/hybrid\_planning.html}); however, it would require the ROS TCP Connector to be updated with support for ROS Actions.

This work also lacks a user study. The purpose of this paper is to introduce the system and its capabilities. Going forward, we plan to conduct a formal user study to compare ERUPT to a 2D computer screen interface.

\section{Conclusion and Future Work}
\label{sec:conclusion}
We present ERUPT, an XR system for interfacing with robot motion planners. Our system enables users to easily modify the robot's environment, maintaining the planning scene as it is edited both in the XR interface and over ROS. In our immersive interface, users can modify collision object geometry and drag to rotate and translate obstacles. Similarly, users can set the trajectory start and goal configuration by directly interacting with the virtual robot. Output paths are visualized to the user so they can evaluate the robot's behavior before executing a path on a physical robot, minimizing risk of undesired motions (such as collisions or proximity to a human operator).
As an open-source, ROS2-integrated platform, ERUPT also provides a foundation for future research in areas including learning from demonstration, teleoperation, and human-in-the-loop planning using XR environments.

In the future, we plan to extend the capabilities of our interface in various ways.
For example, while we support the ability to change planners in our interface, we are unable to change planner specific parameters to customize the chosen planner.
Another capability we would like to support is the ability for users to correct robot trajectories if they are not satisfied with the planner result.
We also plan to add hand-tracking for more intuitive interactions.

\section{Acknowledgments}

This work was supported in part by the IBM-Illinois Discovery Accelerator Institute, the Center
for Networked Intelligent Components and Environments (C-NICE), and by the Illinois-Insper Partnership at the University of Illinois.
Morales is supported in part by Asociación Mexicana de
Cultura A.C.

\bibliographystyle{eg-alpha-doi} 
\bibliography{refs}       




\end{document}